\documentclass{article}

\PassOptionsToPackage{numbers, sort&compress}{natbib}
\usepackage[final,main]{neurips_2026}

\usepackage[utf8]{inputenc} 
\usepackage[T1]{fontenc}    
\usepackage{hyperref}       
\usepackage{url}            
\usepackage{booktabs}       
\usepackage{amsfonts}       
\usepackage{nicefrac}       
\usepackage{microtype}      
\usepackage{xcolor}         

\usepackage{graphicx}
\usepackage{subcaption}
\usepackage{amsmath}
\usepackage{amssymb}
\usepackage{mathtools}
\usepackage{amsthm}
\usepackage{makecell}
\usepackage{placeins}

\usepackage{booktabs}
\usepackage{graphicx}
\usepackage{tikz}
\usetikzlibrary{positioning, shapes.geometric, arrows.meta, calc, fit, backgrounds}
\usepackage[capitalize,noabbrev]{cleveref}

\theoremstyle{plain}

\theoremstyle{definition}

\theoremstyle{remark}

\usepackage[textsize=tiny]{todonotes}

\title{HardwareDx: A Foundational Graph Model \\ System for Datacenter Machine Repairs}

\author{%
  Yuanlin Wen\thanks{Denotes equal contribution. \\ Corresponding author emails: \texttt{\{yuanlinw, elanmark, bperozzi\}@google.com}} \quad
  Elan Markowitz\footnotemark[1] \quad
  Zubo Gu \quad
  Sami Abu-El-Haija\thanks{Work performed while employed at Google} \quad
  Bryan Perozzi \\
  \textbf{Mikhail Galkin} \quad
  \textbf{Ali Parviz}\thanks{Work performed while interning at Google} \quad
  \textbf{Aayush Gupta} \quad
  \textbf{Venkata Chivukula} \quad
  \textbf{Griffin Hadfield} \\
  \textbf{Aravind Akella} \quad
  \textbf{Daniel J. Frame} \quad
  \textbf{Vahab Mirrokni} \quad
  \textbf{Amin Vahdat} \\
  \\
  Google, Mountain View, CA, USA
}

\begin{document}

\maketitle

\begin{abstract}

The rapid scale-up of AI infrastructure has made datacenter reliability a cornerstone of the cloud business economy. Since hardware failures in hyper-scale AI clusters can stall multi-million dollar training workloads, the downtime of individual machines can cause outsized economic losses.
Minimizing machine downtime, therefore, is more important than ever.

To address this, we present \textsc{Hardware Diagnoser (HardwareDx)}, an AI-driven hardware diagnosis system that applies a graph foundation model approach to automated datacenter maintenance. It allows a single model to diagnose many platforms and accelerate onboarding of new hardware products. Deployed at production scale at Google, \textsc{HardwareDx} yields substantial efficiency gains across diverse compute architectures. Real-world evaluations demonstrate that it reduces production downtime by 18.3\% on GPU platforms and 16.1\% on TPUs. Furthermore, it significantly accelerates hardware deployment, cutting quality assurance (QA) time by 28.3\% for GPUs and 25.1\% for TPUs. These results establish \textsc{HardwareDx} as a field-tested solution for maximizing the reliability and cost efficiency of next generation AI fleets.
\end{abstract}

\section{Introduction}
\label{intro}

State-of-the-art foundation models are typically trained using thousands to hundreds-of-thousands of interconnected machines \citep{Comanici2025Gemini2P}. 
When a single machine in a synchronized training cluster fails, the entire workload often halts, leaving thousands of other accelerators idle. In 100,000-GPU clusters, where mean time-to-failure is 15-18 minutes \citep{Kokolis2024RevisitingRI, Salpekar2026TrainingLW}, such failures create massive downtime costs. It is natural to consider leveraging models to predict failures.  Indeed, many prior works~\citep{liang2006bluegene,xue2015practise,botezatu2016predicting,gao2020task,das2018desh} build algorithms and models to anticipate emerging failures. Once failures occur, reducing time-to-repair directly translates to equivalent downtime savings and efficiency, unlocking millions in value and improving velocity as researchers are able to iterate more quickly within a fixed compute budget. 

Site-reliability and data-center engineers aim to mitigate this overhead by quickly detecting and isolating faulty hardware. Historically, repairs were entirely manual. As fleets scaled, engineers moved to standardized actions and simple rules, and eventually to sophisticated score-based rule systems tailored to specific hardware platforms~\citep{yu2026survey,liu2023predicting}. While early machine learning approaches (such as MLPs and decision forests) were explored, they failed to outperform these carefully-engineered rule systems (Appendix~\ref{app:history}). Upon detecting an issue, these systems isolate the faulty hardware and carry out the needed hardware or software repair. While many failures may resolve with just a single component reseat (remove and reconnect), complex comorbidity at the tail-end, such as power anomalies corrupting interconnected high-speed networking switches, can require extensive interventions. To minimize wasted time and unnecessary hardware swaps, the repair rules must be meticulously maintained, leading to fragility over time. Further, the rule set is platform and hardware-model specific~\citep{sridharan2013feng,sridharan2015memory,oliveira2017experimental,meza2015revisiting,vishwanath2010characterizing}. Every new product introduction requires hand-curating a new set of rules for fault diagnosis and repair.

By moving from brittle, manually-maintained rule systems to a transferable graph model, \textsc{HardwareDx} introduces a practical approach for datacenter hardware diagnosis. Specifically, \textsc{HardwareDx} operates as a domain-specific graph foundation model that pretrains across different machine topologies to learn shared structural representations. The pretrained backbone transfers structural knowledge to newly deployed hardware generations with minimal data.
We have a full working implementation of \textsc{HardwareDx} running for 6 months at Google across 50+ ML clusters. Our empirical evaluation shows that \textsc{HardwareDx} classifies each component as (faulty, not-faulty) with >95\% precision at >95\% recall. One single model 
can run inference
across many compute platforms and across multiple tasks for each.

There are several practical benefits over a manually-updated, rule-based system~\citep{he2023unicron,jiang2025l4,wan2025robust,yang2025aidai,cui2025characterizing}. First, the fault detection accuracy is higher (+12.6\%), yielding 18.3\% downtime reduction on GPU platforms and 16.1\% downtime reduction on TPUs. Second, it allows for quick adaptation as new hardware generations are introduced to the market. 
By leveraging LLMs to process log lines~\citep{guo2021logbert,huang2025no,qi2025large}, we eliminate the need for complex parsers and rigid schemas.
This doubles development speed, allowing test engineers to define signatures for {\textasciitilde{}}400 symptoms per platform in just 1 software engineering (SWE) month instead of 2.
We further accelerate development by utilizing transfer learning for new platforms, enabling production-ready deployment up to four months earlier, prior to general availability of the hardware. At our scale, such velocity improvements can unlock significant additional revenue.   

We summarize our contributions and originality as follows. First, we frame the machine repair problem as a node classification task operating over the physical hardware topology. Second, we present a transfer learning approach that adapts generalized representations to new hardware platforms. Third, we introduce a novel combination of learning objectives leveraging both supervised expert labels and unsupervised historical trajectories. Finally, we validate these methods through a large-scale production deployment, establishing, to our knowledge, the first published GNN-based system for AI hardware repairs. \textsc{HardwareDx} is currently deployed for six AI accelerator platforms at Google, including a new TPU platform.

\section{Background}
\label{sec:background}

\definecolor{gblue}{RGB}{66, 133, 244}
\definecolor{gred}{RGB}{234, 67, 53}
\definecolor{gyellow}{RGB}{251, 188, 5}
\definecolor{ggreen}{RGB}{52, 168, 83}
\definecolor{gpurple}{RGB}{103, 58, 183}

\begin{figure}
  \centering
  \begin{tikzpicture}[
    scale=0.8, 
    transform shape,
    node distance=1.5cm and 1.5cm,
    font=\small,
    basebox/.style={rectangle, draw, rounded corners=4pt, minimum width=2.4cm, minimum height=1.2cm, align=center, inner sep=4pt},
    machine/.style={basebox, fill=white, draw=black!70, thick},
    install/.style={basebox, fill=gblue!10, draw=gblue, thick},
    testing/.style={basebox, fill=gyellow!15, draw=gyellow, thick},
    watson/.style={basebox, fill=ggreen!10, draw=ggreen, thick},
    technician/.style={basebox, fill=gred!10, draw=gred, thick},
    check/.style={diamond, draw=gpurple, thick, fill=gpurple!10, aspect=2, inner sep=0pt, minimum width=2.8cm, align=center},
    arrow/.style={-{Stealth[scale=1.0]}, draw=black!60, line width=0.8pt}
  ]

    \node[machine] (new) at (0, 3.5) {New \\ Machine};
    \node[install] (install) at (2.9, 3.5) {Install};
    
    \node[check] (monitor) at (4.6, 1.9) {Monitoring \\ Check};
    
    \node[testing] (test) at (6.6, 3.5) {Testing \\ Suite};
    
    \node[watson] (watson) at (10.1, 3.5) {HardwareDx \\ Diagnoser};
    \node[technician] (repair) at (10.1, 1.9) {Technician \\ Repair \\ Action};

    \draw[arrow] (-1.2, 0.4) -- (12.2, 0.4);
    \node[below, font=\normalsize] at (0.1, 1.0) {Serving Machines};

    \draw[arrow] (new.east) -- (install.west);
    
    \draw[arrow, rounded corners=6pt] (2.3, 0.4) |- (monitor.west);
    
    \draw[arrow] (monitor.south) -- node[fill=white, inner sep=1pt, font=\scriptsize, pos=0.4] {Pass} (monitor.south |- 0, 0.4);
    
    \draw[arrow, rounded corners=6pt] (monitor.north) |- node[fill=white, inner sep=1pt, font=\scriptsize, pos=0.12] {Fail} ([yshift=-10pt]test.west);
    
    \draw[arrow] (install.east) -- (test.west);
    
    \draw[arrow] (test.south) -- node[right, font=\scriptsize, pos=0.25] {Pass} (test.south |- 0, 0.4);
    \draw[arrow] (test.east) -- node[fill=white, inner sep=1pt, font=\scriptsize, pos=0.25] {Fail} (watson.west);
    
    \draw[arrow] (watson.south) -- (repair.north);
    
    \draw[arrow, rounded corners=6pt] (repair.east) -- ++(0.6, 0) |- ($(watson.north) + (0, 0.4)$) -| (test.north);

  \end{tikzpicture}
  \caption{Simplified flow of datacenter operations.}
  \label{fig:system_design}
\end{figure}
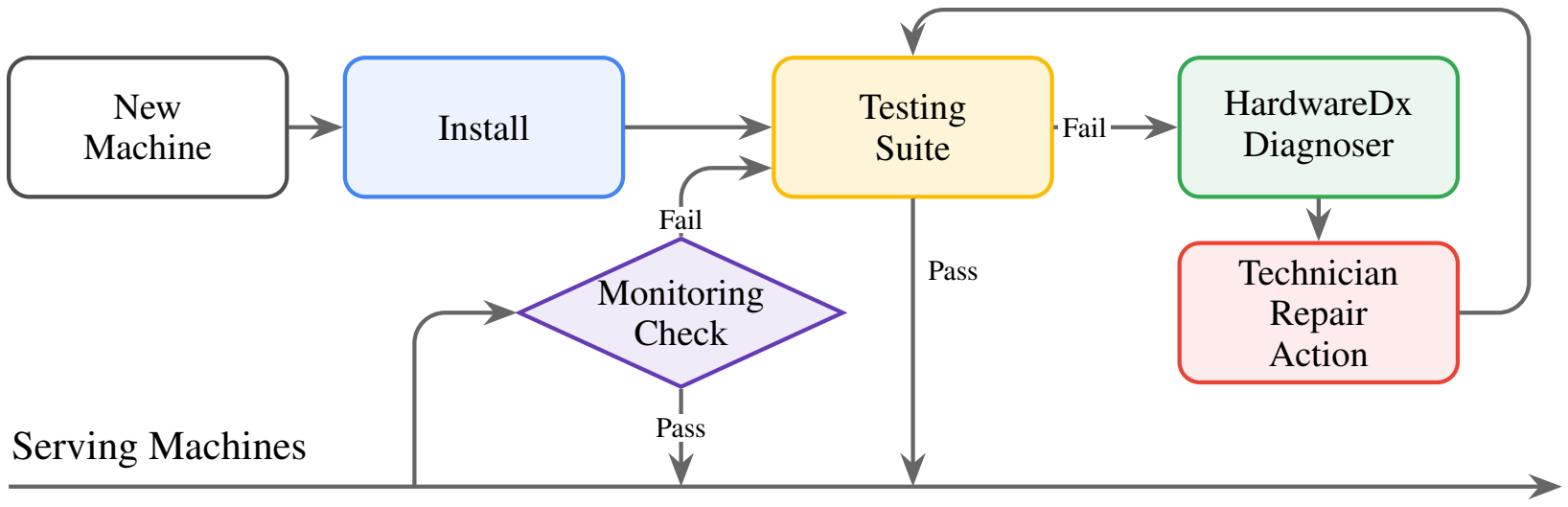

Data center diagnostics and repairs are highly complex operations where a primary concern is uptime for serving machines and rapid quality assurance for new datacenter builds~\citep{borrill2026ghost,levitt2025fine,pu2025errorprism}. From a business perspective, these operations directly impact the bottom line. Uptime for serving machines is a primary driver of cloud revenue, and rapid quality assurance directly corresponds to the server provisioning required to maintain throughput and latency targets for products~\citep{zheng2026sysom,ranganathanempirical}. \textsc{HardwareDx}'s ability to reduce downtime leads directly to increased service availability and reduced capital expenditure on spare capacity. 

IT Quality Assurance (QA) is under even bigger pressure to deliver all capacity to the customer, making long-tail repairs critical~\citep{gao2020task}. In the competitive cloud business, providers vie for the lowest ``Time to Market'', rendering the repair speed of these long-tail failures even more important to ensure timely delivery of massive compute clusters.

Each machine is an expensive grouping of equipment that forms a cohesive unit. For instance, an H100 machine may consist of eight GPUs on a tray, as well as CPUs, memory chips, networking switches, a motherboard, and many types of cables to wire them together~\citep{pattabiraman2026airesim,liu2023predicting,yu2026survey}. When a component fails, the whole machine ceases to function. This means that any workloads relying on that machine must pause as well. If those workloads are large AI training or inference runs, then many other machines could also sit idle given the bulk synchronous nature of these workloads \citep{patel2024100k}. 

A new machine going through setup first goes through an install process where it is assembled and the correct software is installed. Upon passing install and quality assurance, the machine is live and can start serving workloads. However, all machines will eventually fail, thus they need to be monitored. The Monitoring system continuously checks telemetry and error data, and if anything is amiss, it will take the machine offline to undergo testing and repairs. 

Once offline, the testing suite runs a large battery of workloads and stress tests on the machine. This is a time-consuming process taking many hours to complete. The primary output of these tests is a binary indication of whether the machine is healthy or not, as well as data that can be used to determine what repair actions may be needed. If the machine is healthy, it can resume serving. If not healthy, it must be diagnosed. Each machine consists of over 100 Field Replaceable Units (FRUs), any of which could be faulty, and multiple possible repair actions for each one. 

At this point \textsc{HardwareDx} ingests the data from testing and additional information such as the machine topology to output a diagnosis of the faulty component or components. Then, based on the component type repair plan and what prior actions have already been attempted, we prescribe a new repair action or set of actions for the technician to perform. The machine then goes back to the testing suite. If it continues to be unhealthy, this process continues for further iterations. Figure \ref{fig:system_design} gives a high level overview of how the repair systems and operations function.

If \textsc{HardwareDx} fails to make a diagnosis due to service failures, the system falls back to using hand-coded rules. If it exceeds a certain threshold of iterations, the case goes to manual diagnosis by expert technicians. 

\subsection{Foundation Model and Platform Transfer}
\label{sec:transfer}

In a standard supervised learning setup, a new model has to be trained from scratch each time a new platform is released.
However, with limited initial deployment volumes and long delays for diagnostics and failure modes to stabilize, it can take over a year to collect enough data to train a sufficiently strong model. To reduce such long lead times, we introduce foundation models for machine repairs, shifting from siloed, platform-specific training to a generalized model capable of solving machine repairs as soon as the hardware is released.

Rather than training a single model for each specific platform, we instead design our model to be able to handle new platforms with different hardware, tests, and component sets, as long as there is some transferable information. We can then train a model on either a single, data-rich platform or multiple platforms, and then finetune for specific platforms and tasks. 

\section{\textsc{HardwareDx} Model Architecture}
\label{sec:model}

\definecolor{gblue}{HTML}{4285F4}
\definecolor{gred}{HTML}{EA4335}

\newcommand{\drawhetgraph}[2]{
    \begin{scope}[shift={(#1)}]
        \node[main_comp, fill=blue!10] (#2_gpu1) at (-1.5, 0) {GPU 1};
        \node[main_comp, fill=blue!10] (#2_gpu2) at (1.5, 0) {GPU 2};
        \node[main_comp, fill=green!30] (#2_net) at (-1.5, -2.8) {Switch};
        \node[main_comp, fill=orange!30] (#2_pwr) at (1.5, -2.8) {PSU};
        
        \node[sub_comp, fill=blue!10] (#2_g1_np) at (-1.8, -0.8) {};
        \node[sub_comp, fill=blue!10] (#2_g1_pp) at (-1.2, -0.8) {};
        \node[sub_comp, fill=blue!10] (#2_g2_np) at (1.2, -0.8) {};
        \node[sub_comp, fill=blue!10] (#2_g2_pp) at (1.8, -0.8) {};
        \node[sub_comp, fill=green!30] (#2_net_p1) at (-1.8, -2.0) {};
        \node[sub_comp, fill=green!30] (#2_net_p2) at (-1.2, -2.0) {};
        \node[sub_comp, fill=orange!30] (#2_pwr_p1) at (1.2, -2.0) {};
        \node[sub_comp, fill=orange!30] (#2_pwr_p2) at (1.8, -2.0) {};
        
        \node[main_comp, fill=green!15, minimum width=1.0cm, minimum height=0.5cm, font=\tiny, inner sep=1pt] (#2_cab1) at (-1.8, -1.4) {Net\\Cable};
        \node[main_comp, fill=orange!15, minimum width=1.0cm, minimum height=0.5cm, font=\tiny, inner sep=1pt] (#2_cab2) at (0, -1.05) {Power\\Cable};
        \node[main_comp, fill=green!15, minimum width=1.0cm, minimum height=0.5cm, font=\tiny, inner sep=1pt] (#2_cab3) at (0, -1.75) {Net\\Cable};
        \node[main_comp, fill=orange!15, minimum width=1.0cm, minimum height=0.5cm, font=\tiny, inner sep=1pt] (#2_cab4) at (1.8, -1.4) {Power\\Cable};
        
        \node[diag_signal] (#2_symp) at (-2.4, 0.9) {Timeout};
        \node[diag_signal] (#2_sens) at (-0.6, 0.9) {Temp};
        \node[diag_signal] (#2_log)  at (1.5, 0.9) {Error Log};
        
        \draw[sedge] (#2_symp) -- (#2_gpu1);
        \draw[sedge] (#2_sens) -- (#2_gpu1);
        \draw[sedge] (#2_log) -- (#2_gpu2);
        
        \draw[gedge, dashed] (#2_gpu1) -- (#2_g1_np);
        \draw[gedge, dashed] (#2_gpu1) -- (#2_g1_pp);
        \draw[gedge, dashed] (#2_gpu2) -- (#2_g2_np);
        \draw[gedge, dashed] (#2_gpu2) -- (#2_g2_pp);
        \draw[gedge, dashed] (#2_net) -- (#2_net_p1);
        \draw[gedge, dashed] (#2_net) -- (#2_net_p2);
        \draw[gedge, dashed] (#2_pwr) -- (#2_pwr_p1);
        \draw[gedge, dashed] (#2_pwr) -- (#2_pwr_p2);
        
        \draw[gedge] (#2_g1_np) -- (#2_cab1);
        \draw[gedge] (#2_g1_pp) -- (#2_cab2.west);
        \draw[gedge] (#2_g2_np) -- (#2_cab3.east);
        \draw[gedge] (#2_g2_pp) -- (#2_cab4);
        
        \draw[gedge] (#2_cab1) -- (#2_net_p1);
        \draw[gedge] (#2_cab2.east) -- (#2_pwr_p1);
        \draw[gedge] (#2_cab3.west) -- (#2_net_p2);
        \draw[gedge] (#2_cab4) -- (#2_pwr_p2);
        
        \node[fit=(#2_symp) (#2_net) (#2_log) (#2_pwr), inner sep=0pt] (#2) {};
    \end{scope}
}

\newcommand{\drawunivgraph}[2]{
    \begin{scope}[shift={(#1)}]
        \node[unode] (#2_t1) at (-0.6, 0.8) {};
        \node[unode] (#2_t2) at (0, 0.8) {};
        \node[unode] (#2_t3) at (0.6, 0.8) {};
        \node[unode] (#2_s1) at (-0.9, 0.2) {};
        \node[unode] (#2_s2) at (-0.3, 0.2) {};
        \node[unode] (#2_s3) at (0.3, 0.2) {};
        \node[unode] (#2_s4) at (0.9, 0.2) {};
        \node[unode] (#2_d1) at (-0.6, -0.4) {};
        \node[unode] (#2_d2) at (0.6, -0.4) {};
        \node[unode] (#2_c1) at (0, -0.9) {};
        \draw[gray_line] (#2_t1) -- (#2_s1); \draw[gray_line] (#2_t1) -- (#2_s2);
        \draw[gray_line] (#2_t2) -- (#2_s2); \draw[gray_line] (#2_t2) -- (#2_s3);
        \draw[gray_line] (#2_t3) -- (#2_s3); \draw[gray_line] (#2_t3) -- (#2_s4);
        \draw[gray_line] (#2_s1) -- (#2_d1); \draw[gray_line] (#2_s2) -- (#2_d1);
        \draw[gray_line] (#2_s3) -- (#2_d2); \draw[gray_line] (#2_s4) -- (#2_d2);
        \draw[gray_line] (#2_d1) -- (#2_c1); \draw[gray_line] (#2_d2) -- (#2_c1);
        \node[fit=(#2_t1) (#2_s1) (#2_s4) (#2_c1), inner sep=0pt] (#2) {};
    \end{scope}
}

\begin{figure}[]
\centering
\resizebox{\textwidth}{!}{
\begin{tikzpicture}[
    >=stealth,
    basic_box/.style={draw, thick, rectangle, rounded corners=2pt, align=center, minimum height=0.8cm},
    llm_box/.style={basic_box, fill=pink!40, text width=1.5cm},
    enc_box/.style={basic_box, fill=yellow!20, text width=1.8cm},
    core_box/.style={basic_box, fill=yellow!50, text width=3cm, minimum height=1.2cm},
    layer_box/.style={basic_box, fill=orange!15, text width=2cm, minimum height=0.7cm},
    dots_box/.style={basic_box, fill=orange!15, minimum width=1.2cm, minimum height=0.7cm},
    pred_box/.style={draw, thick, rectangle, fill=orange!20, text width=2.2cm, align=center, minimum height=1.0cm},
    unode/.style={draw=black!80, thick, circle, inner sep=2pt, fill=white},
    %
    main_comp/.style={rectangle, draw=black!80, thick, minimum width=1.4cm, minimum height=0.5cm, align=center, rounded corners=1.5mm, font=\scriptsize},
    sub_comp/.style={circle, draw=black!80, thick, minimum size=0.35cm, inner sep=0pt},
    diag_signal/.style={rectangle, draw=red!80, fill=red!10, thick, minimum width=1.1cm, minimum height=0.4cm, align=center, font=\scriptsize},
    %
    gray_line/.style={draw=gray!50, thick},
    arrow/.style={-{Stealth[scale=1.2]}, thick},
    blue_arrow/.style={-{Stealth[scale=1.2]}, thick, gblue},
    red_arrow/.style={-{Stealth[scale=1.2]}, thick, gred},
    gedge/.style={thick, draw=gray!80},
    sedge/.style={thick, draw=red!50, dotted}
]
    \begin{scope}[shift={(-8.0, 1.9)}]
        \draw[thick, draw=gray!50, rounded corners=4pt, fill=gray!5] (-5.6, -1.0) rectangle (5.6, 0.8);
        \node at (0, 0.5) {\scriptsize\textbf{Graph Node \& Edge Legend}};
        
        \node[main_comp, fill=blue!10, minimum width=0.8cm, minimum height=0.4cm, inner sep=0pt] (leg_fru) at (-4.8, 0.0) {};
        \node[right, align=left, font=\footnotesize] at (-4.4, 0.0) {Main Component / FRU};
        
        \node[sub_comp, fill=blue!10] (leg_sub) at (-1.2, 0.0) {};
        \node[right, align=left, font=\footnotesize] at (-1.0, 0.0) {Subcomponent (e.g. Port)};
        
        \node[diag_signal, minimum width=0.8cm, minimum height=0.4cm, inner sep=0pt] (leg_signal) at (2.6, 0.0) {};
        \node[right, align=left, font=\footnotesize] at (3.0, 0.0) {Diagnostic Signal};
        
        \draw[gedge, dashed] (-4.8, -0.6) -- (-4.2, -0.6) node[right, text=black, font=\footnotesize] {Internal Link};
        \draw[gedge] (-1.2, -0.6) -- (-0.6, -0.6) node[right, text=black, font=\footnotesize] {Physical Link};
        \draw[sedge] (2.4, -0.6) -- (3.0, -0.6) node[right, text=black, font=\footnotesize] {Feature Link};
    \end{scope}

    \drawhetgraph{-10.5, -0.5}{g1}

    \node[font=\small\color{gred}, align=left] (text_emb) at (-7.2, -1.0) {Text\\Features};
    \node[llm_box] (llm) at (-5.0, -1.0) {LLM};
    %
    \node[font=\small, align=left] (topo) at (-7.2, -2.25) {Topology};
    %
    \node[font=\small, align=left] (num_feat) at (-7.2, -3.5) {Numeric\\Features};
    \node[enc_box] (enc) at (-5.0, -3.5) {Feature\\Encoder};
    %
    \drawunivgraph{-1.5, -2.45}{ug1}
    \node[font=\small\bfseries\color{gblue}, align=center] at (-1.5, -3.9) {Universal Graph \\ Format};
    %
    \node[core_box] (core_model) at (2.0, -2.5) {\textbf{GNN +} \\ \textbf{Transformer}};
    %
    \begin{scope}[shift={(2.0, 1.5)}]
        \node[layer_box] (gnn1) at (-2.8, 0.8) {GNN};
        \node[layer_box] (trans1) at (-2.8, -0.6) {Transformer};
        \node[dots_box] (dots1) at (0, 0.8) {...};
        \node[dots_box] (dots2) at (0, -0.6) {...};
        \node[layer_box] (gnn2) at (2.8, 0.8) {GNN};
        \node[layer_box] (trans2) at (2.8, -0.6) {Transformer};
        \draw[arrow] (gnn1) -- (trans1);
        \draw[arrow, rounded corners=4pt] (trans1.east) -- ++(0.6,0) |- (dots1.west);
        \draw[arrow] (dots1) -- (dots2);
        \draw[arrow, rounded corners=4pt] (dots2.east) -- ++(0.6,0) |- (gnn2.west);
        \draw[arrow] (gnn2) -- (trans2);
    \end{scope}
    %
    \coordinate (b_mid) at ($(trans1.south west)!0.5!(trans2.south east) + (0, -0.6)$);
    \draw[gray!60, thick, rounded corners=6pt] (trans1.south west) |- (b_mid) -| (trans2.south east);
    \draw[gray!60, thick] (b_mid) -- (core_model.north);
    %
    \drawunivgraph{6.0, -2.45}{ug2}
    %
    \node[pred_box] (p4) at (9.3, -2.05) {};
    \node[pred_box] (p3) at (9.2, -2.2) {};
    \node[pred_box] (p2) at (9.1, -2.35) {};
    \node[pred_box] (p1) at (9.0, -2.5) {Prediction\\Head (GPU)};
    
    \node[align=center, font=\scriptsize] at (9.15, -3.5) {Per-Task \& \\ Per-Component};
    %
    \draw[red_arrow] (text_emb.east) -- (llm.west);
    \draw[arrow] (num_feat.east) -- (enc.west);
    %
    \draw[red_arrow, shorten >= 4pt] (llm.east) -- (ug1_t1.west);
    \draw[blue_arrow, shorten >= 4pt] (topo.east) -- (ug1_s1.west);
    \draw[arrow, shorten >= 4pt] (enc.east) -- (ug1_d1.west);
    %
    \draw[blue_arrow] (ug1.east) -- (core_model.west); 
    %
    \draw[blue_arrow] (core_model.east) -- (ug2.west); 
    %
    \draw[arrow] (ug2.east) -- (p1.west);
    %
    \draw[arrow] (p1.east) -- ++(0.8, 0);
    \draw[arrow] (p2.east) -- ++(0.8, 0);
    \draw[arrow] (p3.east) -- ++(0.8, 0);
    \draw[arrow] (p4.east) -- ++(0.8, 0);
\end{tikzpicture}%
}
\caption{\textsc{HardwareDx} Model Architecture. The machine topology and diagnostic signals are represented as a heterogeneous graph. Each node type is encoded according to the features it has and the topology is homogenized to create the \textit{universal graph format}. This graph then goes through the GNN backbone before predictions are made for each task and component type}
\label{fig:dr_watson_combined}
\end{figure}
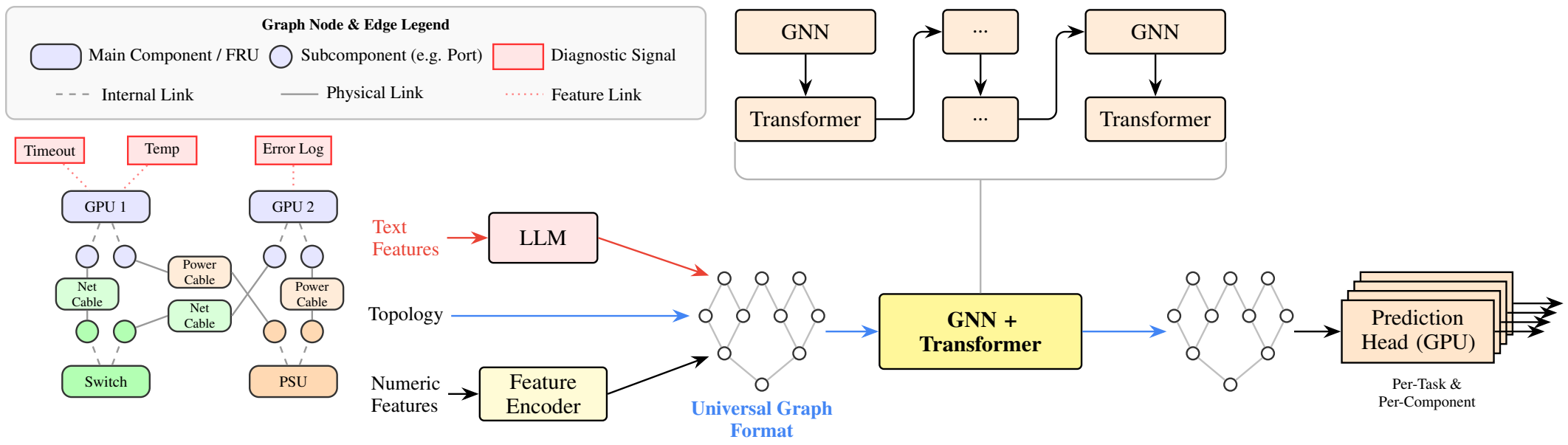

The \textsc{HardwareDx} model centers around a Graph Neural Network (GNN)~\citep{chami2022machine} that uses the physical hardware topology along with other diagnostic signals to predict failing components and repair actions to take. The input includes prior actions, symptoms, sensors, telemetry, and other messages, as well as the machine topology. The model then uses a combination of heterogeneous encoder layers, homogeneous GNN layers, and transformer layers, before predicting a binary faulty label for each Field Replaceable Unit (FRU). An overview of the \textsc{HardwareDx} architecture is illustrated in Figure \ref{fig:dr_watson_combined}.

\subsection{Machine Hardware Topology}

The graph input to \textsc{HardwareDx} is primarily derived from the physical machine topology. Each component is a node in the graph representing either a Field Replaceable Unit (FRU) or a subcomponent of a FRU. For instance, the graph would contain a node for each GPU, which would each connect to subcomponents such as networking ports, power supplies ports, etc. Those subcomponents would connect to other components such as networking cables and power supplies. These components come together to form a heterogeneous graph of nodes of different component types. Figure \ref{fig:dr_watson_combined} (left) illustrates a mock heterogeneous graph representation of this topology.
Each machine is represented as a single, self-contained graph. Different machines are not connected, as hardware topologies and physical faults are isolated at the individual chassis/tray boundary. This also keeps graphs small and inference efficient. Each platform generally has a single physical topology, with occasionally more than one hardware variant, while the attached feature nodes (Section~\ref{sec:features}) vary from graph to graph in both structure and value. Appendix~\ref{app:graph_construction} formally defines the graph.

\subsection{Features: Sensors, Symptoms, Diagnostics}
\label{sec:features}

In addition to the machine topology, the model input also consists of various signals generated during stress testing. These signals include: \textbf{Telemetry Data} such as interconnect bandwidths, device counts, or tflops, as well as sensor readings like temperature, power, and fan-speed; \textbf{Diagnostic Symptoms} resulting from failing tests such as \textit{gpu-clock-speed-below-limit} or \textit{timeout-error}; and \textbf{Error Messages} parsed directly from logs. 

These features are then added as nodes to the graph and attached via an edge to the corresponding component that generated them. Each feature node may have multiple actual feature values. For instance, a fan speed reading would become a node of type "\texttt{sensor}" with features \{\texttt{name:`fan speed', unit:`hertz', value:121.3}\}. Features not generated by a specific component are attached to a ``software'' node. Appendix~\ref{app:features} lists all feature node types, their data sources, and the component types, along with feature-importance ablations.

\subsection{Universal Graph Virtualization and Parameter Homogenization}

The initial machine topology and diagnostic state form a highly heterogeneous graph, defined as $\mathcal{G}_H = (\mathcal{V}_H, \mathcal{E}_H, \mathcal{X}_H)$. Here, the nodes $\mathcal{V}_H$ and edges $\mathcal{E}_H$ are partitioned by diverse component and relation types ($\mathcal{T}_v$ and $\mathcal{T}_e$), and $\mathcal{X}_H$ contains varying raw features ranging from floating-point telemetry to unstructured text logs.

Standard heterogeneous GNNs maintain distinct learnable weight matrices for every discrete relation type \citep{yoon2022zero}. Unfortunately this direct formulation scales poorly, resulting in a combinatorial parameter explosion of $\mathcal{O}(|\mathcal{T}_v|^2 \times |\mathcal{T}_e|)$ \citep{Schlichtkrull2017ModelingRD}. In a large-scale datacenter context, this parameter bloat causes severe overfitting on sparse physical connections and results in catastrophic transfer failure when novel hardware introduces uninitialized relation types.

To resolve this bottleneck, we propose a \textit{virtualization function} that maps $\mathcal{G}_H$ into a unified, homogeneous graph $\mathcal{G}_v = (\mathcal{V}_v, \mathcal{E}_v, H)$. This homogenization results in a mirror graph containing a single node set and a single edge set. For each node $i \in \mathcal{V}_H$ of type $\tau(i)$ with features $x_i$, we deploy a type-specific encoder projection head $f_{\tau(i)}$ (such as MLPs for continuous data or lookup tables for discrete data) to project the heterogeneous features into a shared latent space $\mathbb{R}^d$:

$$h_i^{(0)} = f_{\tau(i)}(x_i)$$

Each encoder is customized to the structure of the data for that node-set. For instance, if a "bandwidth" node type has a float feature for the bandwidth speed and a string indicating the flow type (e.g. "CPU-to-GPU"), then the encoder will be 
projecting the features 
into a single unified representation. 

This architecture decouples feature representation from structural message-passing. By restricting the GNN to operate exclusively on the homogenized graph $\mathcal{G}_v$ using shared parameters,
the model acts as a strong structural regularizer. Consequently, \textsc{HardwareDx} learns generalized topological priors that seamlessly transfer to new compute platforms, requiring only the lightweight adaptation of the isolated node encoders in case new node types appear.

\subsection{LLM Embeddings}

Since many of the features are text strings--either parsed from log files or symptom names--it makes sense to encode these features using an LLM. Additionally, between platforms, we see slight variations in symptom names and logging for the same underlying condition. For instance, "gpu-pcie-bus-error" became "nvfdpilot-gpu-pcie-bus-error" in a platform update. A categorical embedding on symptom type would result in no foundational knowledge transfer between these features. However, with an LLM, the resulting embedding has a high similarity (0.89 cosine similarity). We use Gecko-1B-En \citep{lee2024gecko} as our LLM embedding model for all string-based features to facilitate the appropriate semantic understanding from one hardware platform to another. 

\subsection{Graph Neural Networks \& Long Range Dependencies}



GNNs capture the inherent structural elements of diagnostic data. Spatial relationships provide critical diagnostic signals; for instance, multiple similar symptoms occurring across connected GPUs strongly suggest an issue with a shared upstream component. Conversely, a single isolated symptom might implicate a variety of connected devices, such as a faulty interconnect cable, defective ports, or the underlying hardware attached to that cable. Based on experimental evaluation, we selected the Graph Attention Network v2 (GATv2) \citep{Brody2021HowAA} as our primary message-passing layer to model these local structural dependencies.

Standard Message-Passing Neural Networks (MPNNs) can struggle with long-range dependencies due to over-squashing of information \citep{abu2019mixhop,alon2021on,topping2022understanding}. Our model must diagnose failure syndromes spanning up to eight graph hops and integrate signals from disconnected ``software'' nodes with non-physical actions and software symptoms. While various solutions for handling long-range dependencies exist \citep{gravina_adgn, topping2022understanding}, we use a Graph Transformer based solution. Instead of positional embeddings to encode local structure \citep{dwivedi2021generalization,Ying2021DoTR}, we follow \citet{rampavsek2022recipe} in interleaving Transformer layers with local MPNN layers (GATv2) to directly encode local graph topology. This design permits unconstrained, direct attention between any pair of nodes in the graph, elegantly resolving the dual challenges of long-range, multi-hop syndrome propagation and the integration of disconnected nodes. Ultimately, these design choices are based on experimental validation (Section~ \ref{sec:experiments_architecture}).

\subsection{Data Formulation}
\label{sec:data_formulation}

Let $\mathcal{C}$ be the set of system components (FRUs) and $\mathcal{A}$ be the set of possible repair actions, including a designated ``no-action''. We leverage two distinct data sources for training: an unsupervised historical dataset and a supervised expert-labeled dataset.

\textbf{Historical Diagnostic Sequences (Unsupervised):} The raw telemetry data consists of episodes representing a machine's journey through the repair process. An episode is a sequence of failed tests interspersed with repair actions, terminating when a test finally passes. We formulate this as a trajectory:
\begin{equation}
    \tau = (x_0, A_0, x_1, A_1, \dots, x_{T-1}, A_{T-1}, x_T)
\end{equation}
where $x_t$ represents the diagnostic graph and telemetry features at step $t$, and $A_t \in \{0, 1\}^{|\mathcal{C}| \times |\mathcal{A}|}$ is a binary mask indicating the specific actions applied to specific components after test $x_t$. Each row sums to 1 as only one action can be prescribed to each component, and if no action is prescribed then there is a 1 in the "no-action" column. The subsequent test result $y_{t+1} \in \{0, 1\}$ acts as our binary success label, where $y_{t+1} = 0$ for all steps $t < T-1$ (next test fails), and $y_{t+1} = 1$ for the final step $t = T - 1$ (next test passes). This yields a dataset of historical transitions $\mathcal{D}_{\text{hist}} = \{(x_t, A_t, y_{t+1})\}$.

\textbf{Expert-Labeled Diagnostics (Supervised):} We also develop smaller datasets of expert-annotated ground truth faults, denoted as $\mathcal{D}_{\text{expert}} = \{(x_i, f_i)\}$, where $f_i \in \{0, 1\}^{|\mathcal{C}|}$ is a binary vector indicating which components should be considered faulty. This data takes many months to develop for each platform as new failure modes are observed and labeled. See Appendix \ref{app:data_labeling} for details on data collection. 

\subsection{Multi-Task Prediction Heads and Loss Functions}
\label{sec:multitask_heads}

\textsc{HardwareDx} employs a multi-task learning framework operating on the shared graph representations. The model branches into three distinct prediction heads for each platform, each with a tailored loss function.

\textbf{Task 1: Predict Faulty.}
The first head isolates the diagnostic objective, outputting a binary fault prediction for each component, $\hat{P}_{\text{faulty}} \in [0, 1]^{|\mathcal{C}|}$. This is supervised exclusively by the expert-labeled dataset $\mathcal{D}_{\text{expert}}$. We optimize the standard binary cross-entropy (BCE) loss over all components $c \in \mathcal{C}$:
\begin{equation}
    \mathcal{L}_{\text{faulty}} = -\mathbb{E}_{(x_t, f) \sim \mathcal{D}_{\text{expert}}}\left[\frac{1}{|\mathcal{C}|} \sum_{c \in \mathcal{C}} \left[ f_c \log\left(\hat{P}_{\text{faulty}}(c)\right) + (1 - f_c) \log\left(1 - \hat{P}_{\text{faulty}}(c)\right) \right]\right]
\end{equation}

\textbf{Task 2: Predict Action Propensity.}
The second head models the general behavioral prior of interventions by predicting the propensity of specific actions being taken, regardless of their outcome. For this task, we utilize the complete historical dataset $\mathcal{D}_{\text{hist}}$. The model predicts the probability of each action for each component, $\hat{P}_{\text{prop}} \in [0, 1]^{|\mathcal{C}| \times |\mathcal{A}|}$, utilizing a sigmoid activation. The loss is computed as the expected sigmoid cross-entropy over all datapoints against the observed actions taken by the technicians:
\begin{equation}
    \mathcal{L}_{\text{prop}} = - \mathbb{E}_{(x_t, A_t) \sim \mathcal{D}_{\text{hist}}} \left[ \frac{1}{|\mathcal{C}|}\sum_{(c, a) \in \mathcal{C}\times\mathcal{A}} \left( A_{t}^{(c,a)} \log(\hat{P}_{\text{prop}}^{(c,a)}) + (1 - A_{t}^{(c,a)}) \log(1 - \hat{P}_{\text{prop}}^{(c,a)}) \right) \right]
\end{equation}

\textbf{Task 3: Conditional Likelihood of Success.}
The final head evaluates the viability of specific actions by learning from the entirety of the historical data $\mathcal{D}_{\text{hist}}$, encompassing both failed and successful repair attempts. The goal is to estimate the likelihood of passing the next test upon taking a set of actions, $P(y_{t+1}=1 \mid x_t, A_t)$. The model outputs a dense matrix $\hat{P}_{\text{succ}} \in [0, 1]^{|\mathcal{C}| \times |\mathcal{A}|}$, where each entry $\hat{P}_{\text{succ}}^{(c,a)}$ represents the likelihood of success conditioned on applying action $a$ to component $c$. 

During training, we use the observed action matrix $A_t$ as a binary mask to select only the predictions corresponding to the actions actually taken by technicians. The masked predictions are evaluated against the binary next-test outcome $y_{t+1}$ using a sigmoid BCE loss:
\begin{equation}
    \mathcal{L}_{\text{succ}} = \underset{(x_t, A_t, y_{t+1}) \sim \mathcal{D}_{\text{hist}}}{-\Large\mathbb{E}}\left[\frac{1}{\|A_t\|_1} \sum_{(c,a) : A_t^{(c,a)}=1} \left[ y_{t+1} \log\left(\hat{P}_{\text{succ}}^{(c,a)}\right) + (1 - y_{t+1}) \log\left(1 - \hat{P}_{\text{succ}}^{(c,a)}\right) \right]\right]
    \label{eq:L_succ}
\end{equation}

The total loss optimized during training is the sum of these three components:
\begin{equation}
    \mathcal{L}_{\text{total}} = \mathcal{L}_{\text{faulty}} + \mathcal{L}_{\text{succ}} + \mathcal{L}_{\text{prop}}
\end{equation}

\subsection{Inference and Prescriptive Action Selection}
\label{sec:inference}

At inference time, \textsc{HardwareDx} functions as a prescriptive policy, mapping the current machine state $x_t$ to an optimal repair action. We have two approaches for inference. The primary approach leverages Task 1 ($\hat{P}_{\text{faulty}}$) directly. The second approach leverages the outputs of Task 2 ($\hat{P}_{\text{prop}}$) and Task 3 ($\hat{P}_{\text{succ}}$) to generate a recommendation through a gated, heuristic selection strategy. 

\textbf{Faulty-based Inference}

The faulty-based inference is based on Task 1. Under this use-case, for every component type there is a specific action plan (e.g., For CPU faults: Reseat first, Swap second, Swap Motherboard third). We set a threshold, $\tau_{\text{faulty}} = 0.5$, and any component with a predicted likelihood of being faulty greater than this threshold, $\hat{P}_{\text{faulty}} > \tau_{\text{faulty}}$, is considered faulty, and we follow the next recommended action for that component type. 

\textbf{Success-rate Inference}

The second inference strategy follows the approach of offline RL and causal inference methods by combining reward with propensity \citep{rosenbaum1983central, kumar2019stabilizing} to ensure that we allow policy improvement while avoiding an extrapolation penalty. For rare actions for a given scenario (i.e., off-policy actions), the model's estimate of $\hat{P}_{\text{succ}}$ may be unreasonably high \citep{Fujimoto2018OffPolicyDR}. This is further exacerbated by occasional noise in the data (e.g., mis-recorded actions). 

In a physical datacenter, prescribing an action with zero historical propensity (an out-of-distribution action) could result in unnecessary actions, workmanship issues, loss of trust by technicians, or possibly even cause hardware damage (e.g. swapping high-voltage components while not powered down). We closely follow the work of \citet{kumar2019stabilizing} in using propensity as a threshold for selecting a support-set of eligible actions. This enables the policy to diverge from and improve upon the historic distribution while avoiding unreasonable actions (Appendix~ \ref{app:succ_rate_inference}). 

\subsection{Expert Label Collection}

Predicting component-level faults currently necessitates ground-truth binary labels at the Field Replaceable Unit (FRU) level. In practice, such labels are seldom derived automatically from datacenter operational data, as the inherent noise and ambiguity in telemetry preclude the generation of high-fidelity ground truth. Supervised learning in this domain requires domain experts to retroactively audit these sequences to infer the underlying root cause, a process prone to noise and ambiguity. For example, if multiple components are reseated simultaneously to resolve a single failure, attributing the fault to a specific FRU becomes non-trivial.

To eliminate the "human-in-the-loop" bottleneck and accelerate development velocity, our system optimizes for repair action prediction rather than explicit root-cause labeling. By leveraging repair trajectories directly in the training objective, we mitigate the need for manual annotation while capturing the 
many-to-one mapping between actions and hardware recovery. However, to ensure architectural integrity, we still evaluate the model against a curated "golden set" of recent, expert-filtered repair trajectories that represent the most prevalent hardware syndromes observed in production.

\section{Experiments}
\label{sec:experiments}

\begin{figure}
    \centering
    \includegraphics[width=0.8\linewidth]{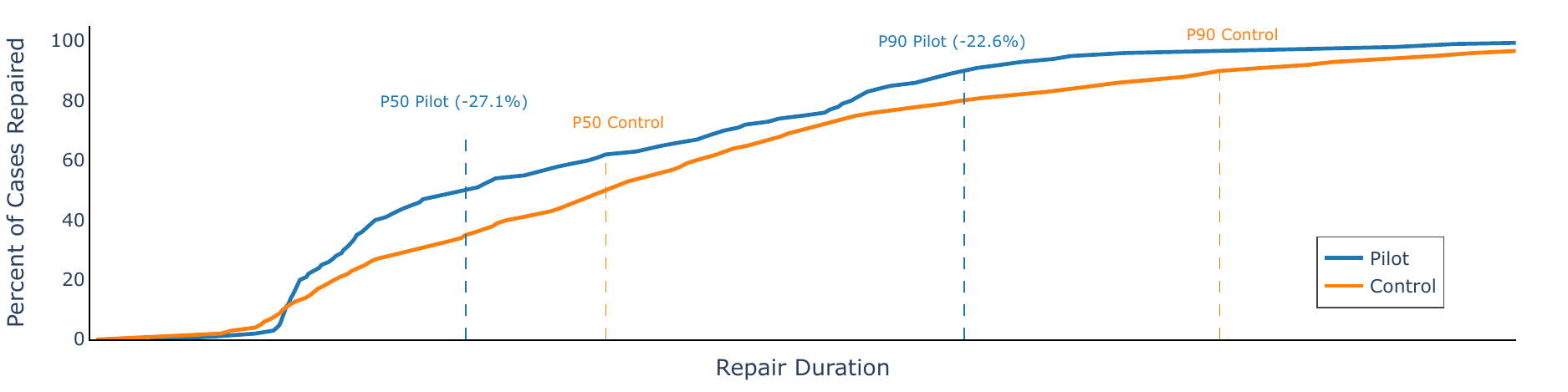}
    \caption{Repair time improvements from deploying \textsc{HardwareDx} on a high-volume TPU platform. \textsc{HardwareDx} improves repair times across the distribution.}
    \label{fig:ab_tpu}
\end{figure}

Our experimental evaluation is designed to answer three primary questions regarding the efficacy of learning-based diagnostic systems in large-scale data centers: 
(1) \textit{Real-World Impact} -- Does a learning-based system outperform meticulously maintained, rule-based expert systems in a live production environment? 
(2) \textit{Sample Efficiency and Transfer Learning} -- Can a foundational model approach significantly reduce the time and data required to support newly introduced hardware platforms?
(3) \textit{Multi-Platform Generalization} -- How does a single, unified foundational model compare to bespoke, single-platform models across diverse GPU and TPU architectures?

\subsection{Real-World Production Impact (A/B Testing)}

\begin{table}[]
    \centering
    \begin{tabular}{lccc}
        \toprule
        \textbf{\begin{tabular}{@{}l@{}}Hardware \\ Platform\end{tabular}} & 
        \textbf{\begin{tabular}{@{}c@{}}Mean TTR \\ Reduction\end{tabular}} & 
        \textbf{\begin{tabular}{@{}c@{}}QA Delivery \\ Reduction\end{tabular}} & 
        \textbf{\begin{tabular}{@{}c@{}}Automated Diagnostic \% \\ (improvement)\end{tabular}} \\
        \midrule
        H100            & 18.3\% $\pm$ 7.6\% & 28.3\% $\pm$ 10.7\% & 93.9\% (+10.7\%) \\
        TPU & 16.1\% $\pm$ 2.6\% & 25.1\% $\pm$ 2.5\% & 82.8\% (+9.9\%) \\
        \bottomrule
    \end{tabular}
    \vspace{0.5em}
    \caption{A/B Production Results: Repair Time Improvements and Automation Coverage. $\pm$ denotes one standard error (Appendix~\ref{app:ab_details}).}  \label{tab:ab_results}  
\end{table}

To evaluate the operational impact of \textsc{HardwareDx}, we conducted a live A/B test in a production environment, comparing its performance against a baseline of the deployed production rules system, which represents a substantial, ongoing engineering investment. Time To Repair (TTR), or downtime, refers to the mean time a machine spends in the test-diagnose-repair loop before returning to serving, excluding install time. Quality Assurance (QA) delivery measures the repair-case percentile needed to reach the contractual cluster-turnover SLO. This rules-based system is the production system that preceded \textsc{HardwareDx}: many hundreds of engineer-maintained rules add ``badness'' scores to the failure modes of each component, and the system prescribes the repair strategy for the highest-scoring component failure mode (Appendix~\ref{app:rules}).

We deployed \textsc{HardwareDx} to predict the faulty components (Task 1) and prescribe initial repair actions across two major hardware categories: the established H100 platform and a newly introduced TPU platform. The A/B analysis is based on 225 H100 repair windows and 2,567 repair windows on the TPU platform. 
Traffic is randomized at the machine level: 20\% of TPU machines and 50\% of H100 machines (a lower-volume pilot) are served by \textsc{HardwareDx}, and the remainder by the rules-based system. Appendix~\ref{app:ab_details} gives further details, including how standard errors are computed.

As shown in Table \ref{tab:ab_results} and Figure \ref{fig:ab_tpu}, \textsc{HardwareDx} yielded substantial improvements across all key operational metrics. For the H100 fleet, \textsc{HardwareDx} reduced the mean Time To Repair (TTR) by 18.3\% and reduced quality-assurance (QA) delivery by 28.3\%. For the TPU platform we observed similar reductions of 16.1\% for mean repair time and a 25.1\% improvement for QA delivery. The large improvement on QA delivery demonstrates the model's ability to handle more complex cases with comorbid failures that typically confound rule-based heuristics and require time-consuming manual diagnosis. Appendix~\ref{app:case_studies} presents two such failure patterns, multiple missing GPUs and comorbid DIMM and PCIe failures, where \textsc{HardwareDx} improves over the rules-based system. These long-tail cases have an outsized impact on deployment time for new datacenter rollouts due to strict QA standards. Finally, the overall 
automated 
diagnostic coverage improved by 10.7\% on H100 (83.1\% to 93.9\%) and 9.9\% on the TPU platform (72.9\% to 82.8\%), proving that ML methods can 
address a broader distribution of failure modes.

\subsection{Transfer Learning and Sample Efficiency}
\begin{figure}[]
    \centering
    \begin{minipage}{0.49\textwidth}
        \centering
        \includegraphics[width=1\linewidth,keepaspectratio]{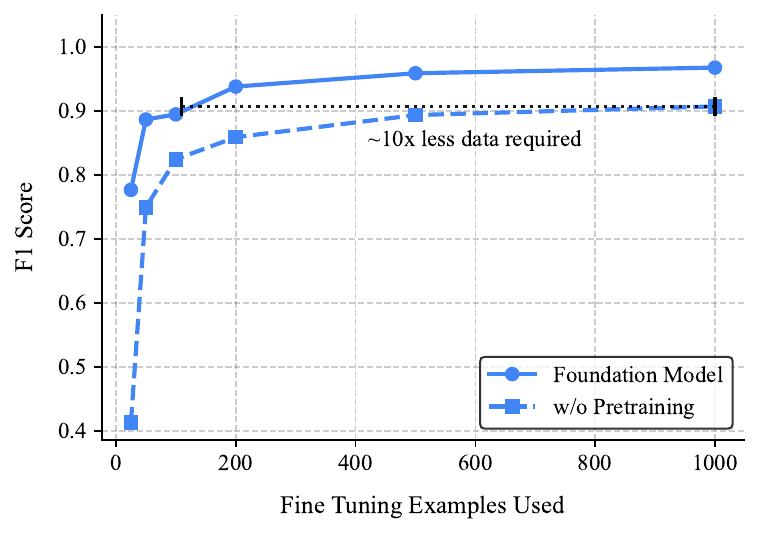}
        \captionof{figure}{Effect of pretraining on transfer learning efficiency. Foundation model pretraining enables onboarding of new platforms with less training data volume. 
        }
        \label{fig:limited_data_finetuning}
    \end{minipage}
    \hfill 
    \begin{minipage}{0.49\textwidth}
        \centering
        \small 
        \begin{tabular}{l c c} 
            \toprule
            & \makecell[c]{\textbf{Job Acc.}} & \makecell[c]{\textbf{Job Recall}} \\
            \midrule
            \makecell[l]{\textbf{Rules}\\ @HW-GA}          & 0.596 & 0.722 \\ \addlinespace
            \makecell[l]{\textbf{\textsc{HardwareDx}-v1}\\ @HW-GA}      & 0.521 & 0.664 \\ \addlinespace
            \makecell[l]{\textbf{\textsc{HardwareDx}}\\ @HW-GA} & 0.722 & 0.776 \\ \midrule
            \makecell[l]{\textbf{\textsc{HardwareDx}-v1}\\ @HW-GA+4mo} & 0.749 & 0.819 \\ \addlinespace
            \makecell[l]{\textbf{\textsc{HardwareDx}}\\ @HW-GA+4mo} & \textbf{0.942} & \textbf{0.969 }\\
            \bottomrule
        \end{tabular}
        \vspace{0.1em}
        \captionof{table}{Performance development relative to general availability of H200 hardware (HW-GA). \textsc{HardwareDx}-v1 is the prior generation of HardwareDx models (no pretraining). 
        }
        \label{tab:performance_timeline}

    \end{minipage}
\end{figure}

A critical bottleneck in data center operations is the "cold-start" problem: when a new compute platform is introduced, it lacks the historical failure data necessary to train a robust diagnostic model from scratch. We evaluated \textsc{HardwareDx}'s ability to leverage foundational knowledge to accelerate this adaptation.

Figure \ref{fig:limited_data_finetuning} demonstrates the sample efficiency and accelerated deployment of our transfer learning approach. We compared a "cold-start" model (trained from scratch on H200 data) against the \textsc{HardwareDx} foundation model, which is pre-trained on H100 telemetry and incrementally fine-tuned. 
\textsc{HardwareDx} provides critical "Day Zero" utility by utilizing virtualized graph representations (Section~\ref{sec:model}) and LLM-based embeddings to recognize failure patterns in new topologies before platform-specific labels exist. Consequently, the model achieves production-grade performance with 10x less data than the cold-start baseline. After 6 months of development, the number of unique observed failure modes was still only half as much as those available on prior platforms. The ability to learn from prior platforms halved the onboarding effort: \textsc{HardwareDx} reached production readiness prior to hardware GA, a significant improvement over the previous generation's four calendar months post-GA requirement. Table \ref{tab:performance_timeline} shows how \textsc{HardwareDx} pulls forward the model readiness relative to hardware GA as measured by job-level accuracy and recall. 
All models in Table~\ref{tab:performance_timeline} are re-evaluated on the same unified held-out set. Historical versions of the rules-based system cannot be replayed, so the Rules@HW-GA row uses the recommendations the production rules made at the time of each repair. Because the rules are continuously updated by a large engineering team, this baseline already includes later rule enhancements, which biases the comparison in favor of the rules (Appendix~\ref{app:rules_baseline}).

\subsection{Foundation Model Multi-Platform Performance}

Next, we evaluate the offline performance of our multi-task foundation model on the component-level fault prediction task. Figure \ref{fig:foundation_vs_single} compares the performance of a foundation model trained on all related platforms vs a model trained on all available data for a single platform.

The results demonstrate that the unified foundational model is highly competitive with, and often superior to, specialized single-platform models. For instance, on the high-volume TPU v7x platform, the foundation model achieves an F1 score of 0.9892, narrowly beating the single model's 0.9877. On newer platforms where data is less ubiquitous, the foundation model's learned structural priors provide a distinct regularization advantage, maintaining high recall without sacrificing precision.

\begin{table}[]
    \centering
    \begin{minipage}{0.48\textwidth}
        \centering
        \includegraphics[width=\linewidth]{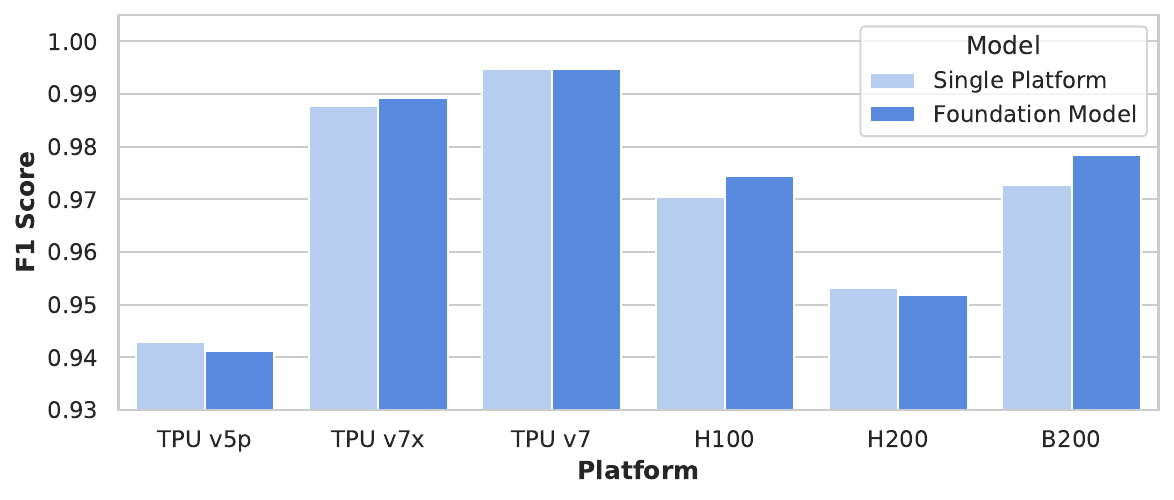}
        \captionof{figure}{F1-score performance comparison of the foundation model vs single platform models. Foundation model performance is generally similar or slightly better than the single platform variants.}
        \label{fig:foundation_vs_single}
    \end{minipage}
    \hfill
    \begin{minipage}{0.48\textwidth}
        \centering
        \resizebox{\linewidth}{!}{%
        \begin{tabular}{lrrr}
            \toprule
            Architecture & F1 Score & Accuracy & Recall \\
            \midrule
            GATV2+Transformer & 0.990 & 0.988 & 0.991 \\
            GIN+Transformer & 0.987 & 0.977 & 0.980 \\
            R-GCN+Transformer$^\dagger$ & 0.988 & 0.980 & 0.985 \\
            GATV2 & 0.985 & 0.978 & 0.986 \\
            GIN & 0.943 & 0.924 & 0.932 \\
            R-GCN$^\dagger$ & 0.974 & 0.953 & 0.962 \\
            \midrule
            Transformer-Only (w/ pos \& path) & 0.929 & 0.846 & 0.848 \\
            Shared Prediction Head & 0.979 & 0.979 & 0.988 \\
            No LLM Features & 0.444 & 0.304 & 0.314 \\
            GBDT (flattened features)$^\ddagger$ & 0.540 & 0.286 & 0.293 \\
            MLP (flattened features)$^\ddagger$ & 0.514 & 0.283 & 0.294 \\
            \bottomrule
        \end{tabular}}
        \vspace{0.5em}
        \caption{Ablation on different architecture choices using TPU v7x for evaluation. Metrics are component-level F1, job-level Accuracy, and job-level Recall. $^\dagger$R-GCN operates on the heterogeneous graph without virtualization and learns relation-specific weights. $^\ddagger$GBDT and MLP are non-GNN baselines.}
        \label{tab:ablation_gnn}

    \end{minipage}
\end{table}

\subsection{Ablations on GNN Architecture}
\label{sec:experiments_architecture}

To justify the architectural design of \textsc{HardwareDx}, particularly the hybrid graph representation discussed in Section~\ref{sec:model}, we evaluate the impact of different GNN layers and the inclusion of the Graph Transformer layer. As shown in Table~\ref{tab:ablation_gnn}, the inclusion of the global attention mechanism (Transformer layer) is critical for handling the long-range dependencies and disconnected software nodes inherent to our diagnostic graphs, consistently improving F1 scores, accuracy, and recall. Furthermore, among the local message-passing layers, GATv2 \citep{Brody2021HowAA} significantly outperforms standard GIN \citep{xu2019powerful}, validating our choice to use attention-based routing for local topology. An R-GCN \citep{Schlichtkrull2017ModelingRD}, which keeps the heterogeneous graph and learns relation-specific weights, does not outperform GATv2 on the virtualized graph either alone (0.974 vs.\ 0.985 F1) or with the Transformer layer (0.988 vs.\ 0.990 F1), so virtualization slightly improves accuracy. 

Table \ref{tab:ablation_gnn} also demonstrates the necessity of structural representation and LLM encodings. Removing the LLM features creates massive performance degradation. Removing the GNN layers and using a Transformer-Only architecture results in a fairly large performance degradation. Note that the Transformer-only model requires positional and device-path encodings to function, which the full GNN model deliberately ignores. We use GNNs because they explicitly provide permutation equivariance and weight sharing across structurally equivalent subgraphs, such as parallel PCIe chains. This allows the model to structurally disambiguate components (e.g., distinguishing whether two failing GPUs reside on the same or different CPU/PCIe chains) without needing additional training examples to transfer a learned pattern from one chain to another. We also analyze a few non-GNN models. We implement Gradient Boosted Decision Trees (GBDT) and a simple Multi-Layer Perceptron (MLP) model using detailed component-level features and aggregated machine-level features. Without the machine topology information, performance significantly degrades (0.54 F1 for GBDT, 0.51 for MLP). Finally, using a shared prediction head across all components slightly degrades performance compared to utilizing individual prediction heads per component. 

\section{Conclusion}

In this paper, we present \textsc{HardwareDx}, a deployed, foundation model system designed to automate hardware diagnostics and prescribe repair actions in large-scale AI datacenters. By shifting away from brittle, manually curated rule-based heuristics, \textsc{HardwareDx} directly addresses the escalating complexity and comorbidity of hardware failures in modern heterogeneous computing fleets. 

Our approach demonstrates the practical viability of applying representation learning to physical infrastructure operations through several key design choices such as graph representation of machine topology, universal graph format, and multi-platform foundation model training. Crucially, \textsc{HardwareDx} effectively mitigates the ``cold-start'' data bottleneck inherent to new hardware deployments. Our live production A/B tests demonstrate substantial operational improvements, including reducing mean time-to-repair by up to 18.3\% and successfully reducing diagnostic QA for new builds by up to 28\%. Furthermore, offline evaluations confirm that our 
transfer learning approach achieves competitive diagnostic accuracy on new platforms up to multiple months earlier than cold-start models.

Ultimately, \textsc{HardwareDx} highlights the immense potential of adapting foundational ML techniques for systems operations. By transitioning from reactive, human-maintained rulesets to proactive, data-efficient models, we pave the way for more resilient, self-healing datacenter ecosystems capable of sustaining the next generation of large-scale AI workloads.

\newpage

\bibliographystyle{plainnat}
\bibliography{drwatson}

\newpage
\appendix

\section{Case Studies}
\label{app:case_studies}

We present two case studies on failure patterns where \textsc{HardwareDx} improved performance over the rules-based system. These failures showcase two primary advantages of GNN-based representation learning: topology awareness and comorbidity handling. Additionally, the learned patterns impact multiple platforms, highlighting how the foundation model accelerates development velocity for new hardware iterations.

\subsection{Case 1: Multiple GPUs Missing}

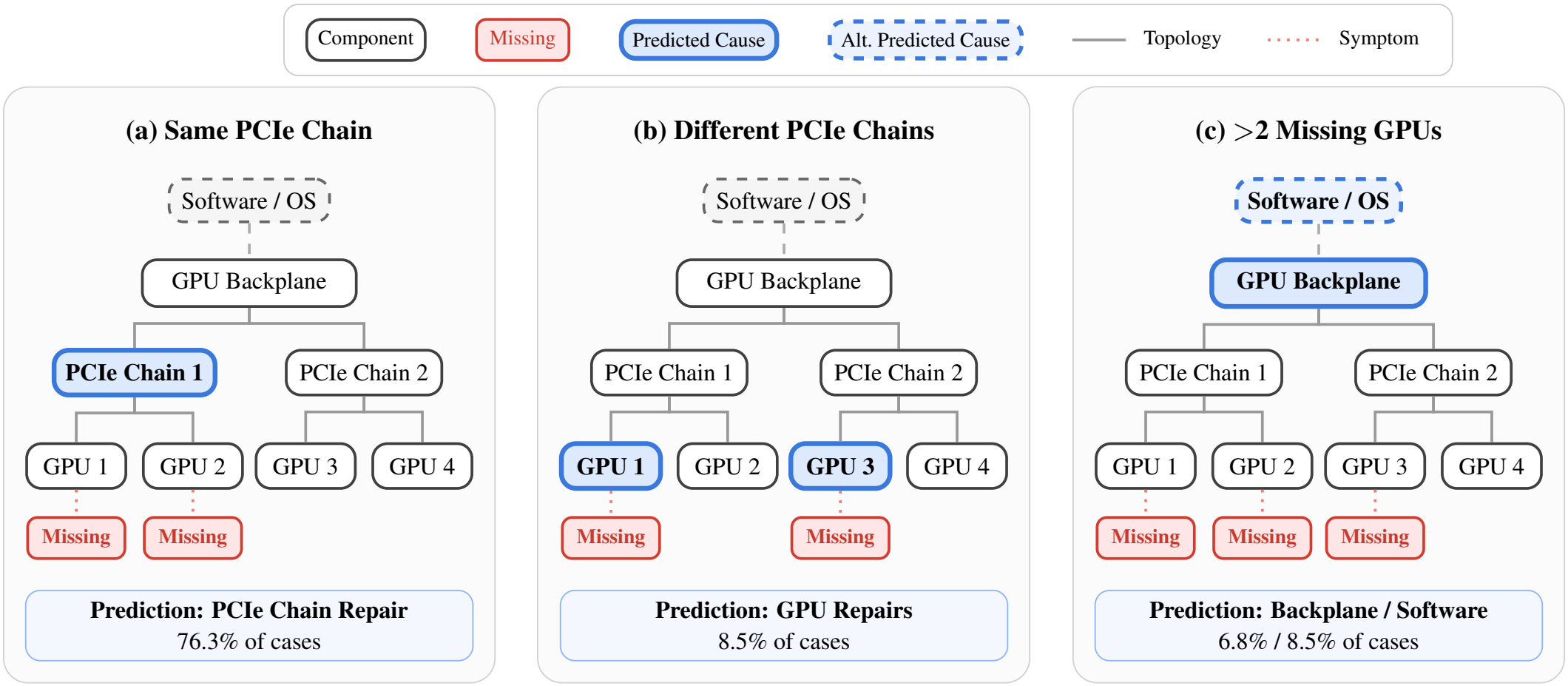
\begin{figure}[htbp]
\centering
\definecolor{gblue}{HTML}{4285F4}
\definecolor{gred}{HTML}{EA4335}
\resizebox{\textwidth}{!}{%
\begin{tikzpicture}[
    hw_node/.style={rectangle, draw=black!75, fill=white, thick, minimum width=1.15cm, minimum height=0.46cm, align=center, rounded corners=1.5mm, font=\scriptsize},
    sw_node/.style={rectangle, draw=black!60, fill=gray!8, thick, dashed, minimum width=1.65cm, minimum height=0.44cm, align=center, rounded corners=1.5mm, font=\scriptsize},
    pcie_node/.style={rectangle, draw=black!75, fill=white, thick, minimum width=1.6cm, minimum height=0.46cm, align=center, rounded corners=1.5mm, font=\scriptsize},
    bp_node/.style={rectangle, draw=black!75, fill=white, thick, minimum width=2.2cm, minimum height=0.48cm, align=center, rounded corners=1.5mm, font=\scriptsize},
    gpu_node/.style={rectangle, draw=black!75, fill=white, thick, minimum width=1.02cm, minimum height=0.46cm, align=center, rounded corners=1.5mm, font=\scriptsize},
    pred_node/.style={draw=gblue!90!black, fill=gblue!18, line width=1.4pt, text=black, font=\scriptsize\bfseries},
    pred_alt_node/.style={draw=gblue!90!black, fill=gblue!10, line width=1.2pt, dashed, text=black, font=\scriptsize\bfseries},
    symptom_node/.style={rectangle, draw=gred!90!black, fill=red!10, thick, minimum width=1.0cm, minimum height=0.38cm, align=center, rounded corners=1mm, font=\tiny\bfseries, text=gred!85!black},
    gedge/.style={thick, draw=gray!80},
    sedge/.style={thick, draw=gred!70, dotted},
    ledge/.style={thick, draw=gray!65, dashed},
    panel_box/.style={draw=gray!35, fill=gray!3, rounded corners=2.5mm, inner sep=0.22cm},
    outcome_box/.style={rectangle, draw=gblue!60, fill=gblue!6, rounded corners=1.5mm, minimum width=4.6cm, minimum height=0.72cm, align=center, font=\scriptsize}
]
\node[hw_node, minimum width=1.05cm, minimum height=0.36cm, font=\tiny] (leg_hw) at (-4.30, 5.10) {Component};
\node[symptom_node, minimum width=0.95cm, minimum height=0.34cm, font=\tiny, right=0.50cm of leg_hw] (leg_sym) {Missing};
\node[hw_node, pred_node, minimum width=1.25cm, minimum height=0.36cm, font=\tiny, right=0.50cm of leg_sym] (leg_pred) {Predicted Cause};
\node[hw_node, pred_alt_node, minimum width=1.35cm, minimum height=0.36cm, font=\tiny, right=0.50cm of leg_pred] (leg_alt) {Alt.\ Predicted Cause};
\coordinate (leg_e1_l) at ($(leg_alt.east)+(0.50,0)$);
\coordinate (leg_e1_r) at ($(leg_e1_l)+(0.55,0)$);
\draw[gedge] (leg_e1_l) -- (leg_e1_r);
\node[font=\tiny, anchor=west] (leg_e1_t) at ($(leg_e1_r)+(0.06,0)$) {Topology};
\coordinate (leg_e2_l) at ($(leg_e1_t.east)+(0.35,0)$);
\coordinate (leg_e2_r) at ($(leg_e2_l)+(0.55,0)$);
\draw[sedge] (leg_e2_l) -- (leg_e2_r);
\node[font=\tiny, anchor=west] (leg_e2_t) at ($(leg_e2_r)+(0.06,0)$) {Symptom};
\node[draw=gray!40, fill=white, rounded corners=1.5mm, inner xsep=0.22cm, inner ysep=0.14cm, fit=(leg_hw)(leg_sym)(leg_pred)(leg_alt)(leg_e2_t)] (legend_box) {};
\node[hw_node, minimum width=1.05cm, minimum height=0.36cm, font=\tiny] at (leg_hw) {Component};
\node[symptom_node, minimum width=0.95cm, minimum height=0.34cm, font=\tiny] at (leg_sym) {Missing};
\node[hw_node, pred_node, minimum width=1.25cm, minimum height=0.36cm, font=\tiny] at (leg_pred) {Predicted Cause};
\node[hw_node, pred_alt_node, minimum width=1.35cm, minimum height=0.36cm, font=\tiny] at (leg_alt) {Alt.\ Predicted Cause};
\draw[gedge] (leg_e1_l) -- (leg_e1_r);
\node[font=\tiny, anchor=west] at ($(leg_e1_r)+(0.06,0)$) {Topology};
\draw[sedge] (leg_e2_l) -- (leg_e2_r);
\node[font=\tiny, anchor=west] at ($(leg_e2_r)+(0.06,0)$) {Symptom};
\begin{scope}[shift={(-5.5, 0)}]
    \node[font=\footnotesize\bfseries] (a_title) at (0, 4.15) {(a) Same PCIe Chain};
    \node[sw_node] (a_sw) at (0, 3.45) {Software / OS};
    \node[bp_node] (a_bp) at (0, 2.60) {GPU Backplane};
    \node[pcie_node, pred_node] (a_p1) at (-1.18, 1.68) {PCIe Chain 1};
    \node[pcie_node] (a_p2) at (1.18, 1.68) {PCIe Chain 2};
    \node[gpu_node] (a_g1) at (-1.78, 0.72) {GPU 1};
    \node[gpu_node] (a_g2) at (-0.58, 0.72) {GPU 2};
    \node[gpu_node] (a_g3) at (0.58, 0.72) {GPU 3};
    \node[gpu_node] (a_g4) at (1.78, 0.72) {GPU 4};
    \node[symptom_node] (a_s1) at (-1.78, -0.02) {Missing};
    \node[symptom_node] (a_s2) at (-0.58, -0.02) {Missing};
    \draw[ledge] (a_sw) -- (a_bp);
    \draw[gedge] (a_bp.south) -- ++(0,-0.16) -| (a_p1.north);
    \draw[gedge] (a_bp.south) -- ++(0,-0.16) -| (a_p2.north);
    \draw[gedge] (a_p1.south) -- ++(0,-0.16) -| (a_g1.north);
    \draw[gedge] (a_p1.south) -- ++(0,-0.16) -| (a_g2.north);
    \draw[gedge] (a_p2.south) -- ++(0,-0.16) -| (a_g3.north);
    \draw[gedge] (a_p2.south) -- ++(0,-0.16) -| (a_g4.north);
    \draw[sedge] (a_g1) -- (a_s1);
    \draw[sedge] (a_g2) -- (a_s2);
    \node[outcome_box] (a_out) at (0, -0.92) {\textbf{Prediction: PCIe Chain Repair}\\[1pt]76.3\% of cases};
\end{scope}
\begin{scope}[shift={(0, 0)}]
    \node[font=\footnotesize\bfseries] (b_title) at (0, 4.15) {(b) Different PCIe Chains};
    \node[sw_node] (b_sw) at (0, 3.45) {Software / OS};
    \node[bp_node] (b_bp) at (0, 2.60) {GPU Backplane};
    \node[pcie_node] (b_p1) at (-1.18, 1.68) {PCIe Chain 1};
    \node[pcie_node] (b_p2) at (1.18, 1.68) {PCIe Chain 2};
    \node[gpu_node, pred_node] (b_g1) at (-1.78, 0.72) {GPU 1};
    \node[gpu_node] (b_g2) at (-0.58, 0.72) {GPU 2};
    \node[gpu_node, pred_node] (b_g3) at (0.58, 0.72) {GPU 3};
    \node[gpu_node] (b_g4) at (1.78, 0.72) {GPU 4};
    \node[symptom_node] (b_s1) at (-1.78, -0.02) {Missing};
    \node[symptom_node] (b_s3) at (0.58, -0.02) {Missing};
    \draw[ledge] (b_sw) -- (b_bp);
    \draw[gedge] (b_bp.south) -- ++(0,-0.16) -| (b_p1.north);
    \draw[gedge] (b_bp.south) -- ++(0,-0.16) -| (b_p2.north);
    \draw[gedge] (b_p1.south) -- ++(0,-0.16) -| (b_g1.north);
    \draw[gedge] (b_p1.south) -- ++(0,-0.16) -| (b_g2.north);
    \draw[gedge] (b_p2.south) -- ++(0,-0.16) -| (b_g3.north);
    \draw[gedge] (b_p2.south) -- ++(0,-0.16) -| (b_g4.north);
    \draw[sedge] (b_g1) -- (b_s1);
    \draw[sedge] (b_g3) -- (b_s3);
    \node[outcome_box] (b_out) at (0, -0.92) {\textbf{Prediction: GPU Repairs}\\[1pt]8.5\% of cases};
\end{scope}
\begin{scope}[shift={(5.5, 0)}]
    \node[font=\footnotesize\bfseries] (c_title) at (0, 4.15) {(c) $>$2 Missing GPUs};
    \node[sw_node, pred_alt_node] (c_sw) at (0, 3.45) {Software / OS};
    \node[bp_node, pred_node] (c_bp) at (0, 2.60) {GPU Backplane};
    \node[pcie_node] (c_p1) at (-1.18, 1.68) {PCIe Chain 1};
    \node[pcie_node] (c_p2) at (1.18, 1.68) {PCIe Chain 2};
    \node[gpu_node] (c_g1) at (-1.78, 0.72) {GPU 1};
    \node[gpu_node] (c_g2) at (-0.58, 0.72) {GPU 2};
    \node[gpu_node] (c_g3) at (0.58, 0.72) {GPU 3};
    \node[gpu_node] (c_g4) at (1.78, 0.72) {GPU 4};
    \node[symptom_node] (c_s1) at (-1.78, -0.02) {Missing};
    \node[symptom_node] (c_s2) at (-0.58, -0.02) {Missing};
    \node[symptom_node] (c_s3) at (0.58, -0.02) {Missing};
    \draw[ledge] (c_sw) -- (c_bp);
    \draw[gedge] (c_bp.south) -- ++(0,-0.16) -| (c_p1.north);
    \draw[gedge] (c_bp.south) -- ++(0,-0.16) -| (c_p2.north);
    \draw[gedge] (c_p1.south) -- ++(0,-0.16) -| (c_g1.north);
    \draw[gedge] (c_p1.south) -- ++(0,-0.16) -| (c_g2.north);
    \draw[gedge] (c_p2.south) -- ++(0,-0.16) -| (c_g3.north);
    \draw[gedge] (c_p2.south) -- ++(0,-0.16) -| (c_g4.north);
    \draw[sedge] (c_g1) -- (c_s1);
    \draw[sedge] (c_g2) -- (c_s2);
    \draw[sedge] (c_g3) -- (c_s3);
    \node[outcome_box] (c_out) at (0, -0.92) {\textbf{Prediction: Backplane / Software}\\[1pt]6.8\% / 8.5\% of cases};
\end{scope}
\begin{pgfonlayer}{background}
    \node[panel_box, fit=(a_title)(a_g1)(a_g4)(a_out)] (a_bg) {};
    \node[panel_box, fit=(b_title)(b_g1)(b_g4)(b_out)] (b_bg) {};
    \node[panel_box, fit=(c_title)(c_g1)(c_g4)(c_out)] (c_bg) {};
\end{pgfonlayer}
\end{tikzpicture}%
}
\caption{Illustration of topology-aware diagnosis in Case 1 (Multiple GPUs Missing). Identical \texttt{Missing} telemetry signals on GPUs lead \textsc{HardwareDx} to distinct repair predictions based on the underlying PCIe hierarchy: (a) two missing GPUs sharing a PCIe chain implicate the shared PCIe chain (76.3\%), (b) two missing GPUs across separate PCIe chains localize the fault to the individual GPUs (8.5\%), and (c) more than two missing GPUs across chains escalate to the GPU backplane (6.8\%) or software (8.5\%).}
\label{fig:case1_missing_gpus}
\end{figure}

This case study highlights the advantage of topology awareness through GNNs. It concerns machines reporting multiple GPUs as missing or unreachable. While a single missing GPU is usually a straightforward indicator of a faulty GPU, cases involving multiple missing GPUs are more complex and challenging for the rules-based system. 

Our findings demonstrate high interdependence on the structural topology around the afflicted GPUs (Figure~\ref{fig:case1_missing_gpus}). When two GPUs are missing, two distinct scenarios exist based on the PCIe chains.
If the same PCIe chain controls both missing GPUs, the model correctly predicts that the PCIe chain requires repair. Conversely, if different PCIe chains control the GPUs, the model learns that GPU repairs are necessary, indicating that either the GPUs or the backplane is faulty. When more than two GPUs are missing, the model typically infers that the backplane is faulty or software issues are present. Table \ref{tab:multiple_gpu_missing} presents the breakdown of recommended repairs.

Because this diagnosis requires attention to the machine's structural layout, it is difficult for rules-based systems to detect these patterns. However, \textsc{HardwareDx} naturally learns to distinguish between them.

\begin{table}[h]
    \centering
    \begin{tabular}{lc}
        \toprule
        \textbf{Category} & \textbf{Percentage} \\
        \midrule
        GPU Backplane & 6.8\% \\
        GPU repairs & 8.5\% \\
        PCIe repairs & 76.3\% \\
        Software & 8.5\% \\
        \bottomrule
    \end{tabular}
    \vspace{0.5em}
    \caption{Breakdown of recommendations by \textsc{HardwareDx} when encountering multiple missing GPUs. Cases are categorized by the most involved repair recommended (Backplane > GPU > PCIe > Software). Cases with comorbid symptoms outside these categories (9.7\%) were excluded.}
    \label{tab:multiple_gpu_missing}
\end{table}

\subsection{Case 2: DIMM Errors with PCIe Low Throughput}

\textsc{HardwareDx} excels at handling comorbid failures. This case examines a scenario involving both excessive DIMM errors and low PCIe throughput. Through \textsc{HardwareDx}, we determined that the DIMM was the ultimate root cause of the issue.

Individually, the rules handle these failures gracefully: excessive DIMM errors trigger DIMM repairs, and low PCIe throughput triggers PCIe chain repairs. However, when both modes appear, the rules trigger both responses. This leads to either extraneous part swaps or increased repair iterations if PCIe repairs are prioritized over the DIMM.

In this scenario, a definitive correct answer exists. The model learns to prioritize DIMM repairs when these symptoms are comorbid. Further analysis revealed that faulty DIMMs with reduced capacities were throttling bandwidth, thereby causing the PCIe errors. This learned pattern facilitated additional diagnostic development and rule adjustments, affecting 2.9\% of all PCIe low-throughput cases on ML platforms.

\section{Related Work}
\label{sec:related}

\textbf{Log Analysis and Diagnostics:} Existing literature on general diagnostics addresses log analysis and anomaly detection through various methodologies \citep{Chandola2009AnomalyDA, Huang2024DemystifyingAE, Lin2016LogCB}. Initial approaches utilized statistical extraction techniques to process system logs \citep{lin2020fast}. Subsequent research introduced natural language processing methods, including self-supervised training for anomaly detection (LogBERT, \citep{guo2021logbert}) and prompt-based few-shot learning (LogPPT, \citep{le2023log}). Recent studies have evaluated the application of large language models (LLMs) for log-based fault diagnosis. For instance, \citet{sun2025accurate} applied fine-tuned LLMs alongside fault-oriented log summaries to process extended log sequences. In the context of distributed training, \citet{jiang2025l4} analyzed spatial and temporal log patterns to diagnose failures in large-scale environments. Additionally, the representation of unstructured telemetry data has been explored using distilled text embeddings, such as those evaluated by \citet{lee2024gecko}. In the context of data center reliability, \citet{Li2021AutomatedIHA} developed an automated intelligent healing framework that suggests software remedies or hardware repair required, but does not suggest the specific hardware repairs. 

\textbf{Graph Neural Networks for Diagnostics:} The structural dependencies within physical topologies have been modeled using graph neural networks (GNNs) \citep{kipf2016semi}. To address complex fault propagation, various specialized GNN architectures have been evaluated. For instance, \citet{Yang2025AnIF} integrated a GNN variant alongside deep reinforcement learning and large language models for self-healing mechanisms in cloud AI software. Similar to \citet{Li2021AutomatedIHA}, their work focused on software remedies rather than actual hardware repairs. In edge computing contexts, \citet{su2026st} proposed a spatio-temporal graph propagation model (ST-GraphRCA) to trace root causes by analyzing anomaly flows across connected nodes. Outside of the diagnostic regimes, other works have focused on GNNs for chip design \citep{Mirhoseini2021AGP}.

\textbf{Prescriptive Maintenance and Transfer Learning:} Research on maintenance scheduling includes the development of prescriptive maintenance frameworks \citep{Giacotto2025Prescriptive}. For example,  \citet{eldele2025unifault} use time-series foundation models to predict bearing failures in rotating machinery. Within this domain, offline reinforcement learning (RL) has been applied to historical datasets to optimize maintenance policies \citep{khorasgani2021offline, Levine2020OfflineRL}. To manage out-of-distribution actions, several methods incorporate propensity scoring and behavioral cloning to constrain the action space \citep{Fujimoto2018OffPolicyDR, kumar2019stabilizing, rosenbaum1983central}. 

\textbf{Graph Foundation Models (GFMs):}
Graph representation learning has been experiencing a "foundation model" shift away from task-specific training toward models that generalize across diverse graph structures. This field has seen significant recent activity \citep{Wang2025GraphFM, wang2024gft}.

Early works like DeepWalk \citep{Perozzi2014DeepWalkOL} and Node2Vec \citep{Grover2016node2vecSF} laid the groundwork for learning structural graph embeddings that can be applied to new tasks. As the field turned towards GNNs \citep{kipf2016semi}, inductive reasoning became a higher priority with works like GraphSAGE \citep{Hamilton2017InductiveRL} focusing on generalization to new graph structures. Work developed further on generalizing graphs to new nodes \citep{Chen2023GeneralizingTU, Teru2019InductiveRP, Daza2020InductiveER, Markowitz2022StATIKSA}, new relations, and eventually entirely new graph schemas \citep{Galkin2023TowardsFM, Wang2024TowardsGF}. 

In the modern era, additional work has looked more specifically at using language models with graphs to create foundation models \citep{chen2024text}. Examples include One-for-All \citep{Liu2023OneFA} and UniGraph \cite{He2024UniGraphLA}, which use LLMs to allow GNNs to bridge different graph domains and schemas. GraphToken \citep{Perozzi2024LetYG} reverses this paradigm, using a GNN to encode graphs as tokens for LLM to process. \textsc{HardwareDx} incorporates language models for encoding features which supports learning on new symptoms and components.

\section{Historical Context of Repair Diagnosis}
\label{app:history}

Early on, hardware debugging at Google was a manual process without standardized directives, in which engineers and technicians decided which actions to take. As the fleet scaled, technician actions were standardized and simple automated rules were developed. These systems were long considered adequate, as individual components were inexpensive, diagnostic signals were more straightforward, and falling back to manual diagnosis was not too costly when downtime was less critical. With AI datacenters this is no longer the case. Components are very expensive, the deeper connectivity of machines leads to more complicated failures, and repair time is increasingly costly. The rules have evolved into a more complex score-based system (Appendix~\ref{app:rules}), but it has not kept pace with the rate of new platform introductions. Our early ML experiments with standard approaches, such as featurized MLPs and decision forests, did not reach the performance necessary to improve on the rules-based system, which led to the GNN-based approach presented here.

\section{Limitations and Future Work}
\label{sec:limitations}

\textbf{Reproducibility and Data Constraints:} Direct replication of our empirical results is constrained by the proprietary nature of datacenter hardware telemetry, fault logs, and maintenance records, which cannot be publicly released due to security and confidentiality requirements. Additionally, live production A/B tests cannot be replayed historically. To maximize conceptual and algorithmic reproducibility, we provide comprehensive architectural specifications (Section~\ref{sec:model}), formal graph schemas and edge categories (Appendix~\ref{app:graph_construction}), feature node keys and component types (Appendix~\ref{app:features}), mathematical formulations of inference heads (Appendix~\ref{app:succ_rate_inference}), and extensive ablation studies across multiple architectures and baselines (Table~\ref{tab:ablation_gnn}). These details are intended to enable researchers and infrastructure operators to implement graph-based diagnostic models within their own compute environments.

\textbf{Baseline Comparison Limitations:} Because production rule engines cannot be backtested against historical machine states without active re-execution, our longitudinal comparison in Table~\ref{tab:performance_timeline} evaluates the production rules available at HW-GA rather than a retrospectively modified Rules@HW-GA+4mo baseline, as discussed in Appendix~\ref{app:rules}.

\textbf{Data Limitations:} Our formulation for task 3 suffers from issues of temporal credit assignment \citep{Sutton1984TemporalCA, Pignatelli2023ASO}. Specifically, our data formulation assigns a success label of 0 to all intermediate steps in a repair trajectory, assigning 1 only to the final passing step. This risks penalizing necessary partial repairs in co-morbid failure scenarios. Furthermore, Equation \ref{eq:L_succ} evaluates the global test outcome against a binary mask of actions, assigning the exact same success signal to all concurrent actions (e.g., a necessary action on component A and an unnecessary action on component B), which introduces causal confounding. This can further obfuscate interpretation as looking at the joint predictions of two correlated actions can imply a different interpretation than looking at the joint predictions for two independent actions. 

Future work will incorporate advances in Credit Assignment to better assign labels (rewards) to specific actions \citep{Alipov2021TowardsPC,Qu2024LatentRL}.

\textbf{Latency:} Serving ML models such as ours introduces additional services and computation, leading to higher latency. However, given that testing can take many hours and wait-times for technicians are even higher, the diagnosis latency is not a consideration for our system. 

\textbf{Scope of Foundation Model and Trajectory Supervision:} In this work, "foundation model" refers to a domain-specific graph architecture that pretrains across heterogeneous datacenter machine topologies to learn shared structural priors. While this enables effective cross-platform transfer within internal fleets, \textsc{HardwareDx} is not an open general-domain model and does not operate in a fully self-supervised manner. Downstream fine-tuning and deployment still depend on platform-specific topology definitions, telemetry schemas, and curated technician repair trajectories. Evaluating transferability on open or synthetic infrastructure benchmarks without proprietary telemetry remains future work.

\textbf{Fully-Unsupervised Systems:} While \textsc{HardwareDx} greatly improves efficiency via foundation model and transfer learning, there is still the requirement for some amount of labeled data and manual effort. This is a limitation we are actively addressing. Appendix~\ref{app:succ_rate_inference} presents a label-free inference approach and Appendix~\ref{app:sim_eval_succ_rate} presents initial results. Future work will further develop our ability to transition to fully unsupervised foundation model systems.

\section{Graph Construction}
\label{app:graph_construction}

\textbf{Heterogeneous graph.} We use standard approaches to encoding heterogeneous graphs. A heterogeneous graph is defined by a collection of node sets and a collection of edge sets, where each edge set has an associated source node set and target node set.

\textbf{Node sets.} Node sets include different hardware component types (e.g., \texttt{cpu}, \texttt{gpu}, \texttt{pcie\_cable}), different telemetry readings (e.g., \texttt{sensor}, \texttt{bandwidth}), errors (e.g., \texttt{dimm\_errors}, \texttt{oob\_errors}), and diagnostic symptoms (\texttt{symptoms}). Appendix~\ref{app:features} lists the feature node types and component types in full.

\textbf{Edge sets.} Edge sets fall into three categories. Physical edge sets connect different component types (e.g., \texttt{cpu}-to-\texttt{pcie\_cable} represents a PCIe cable plugged into a CPU connector). Internal edge sets capture physical or logical connections within a component (e.g., the \texttt{cpu}-to-\texttt{cpu} edge set links a CPU's PCIe port subcomponent to the specific CPU node it belongs to). Feature edge sets connect feature nodes to the component that generated them (e.g., \texttt{sensor}-to-\texttt{cpu} and \texttt{symptom}-to-\texttt{gpu}).

\textbf{Graph scope.} Each machine is represented as a single, self-contained graph. We do not connect different machines in the graph, as hardware topologies and physical faults are isolated at the individual chassis/tray boundary. This keeps the graph size small and inference efficient. Applying a similar approach to inter-node communication hardware is a possible extension of this work. Each platform generally has a single physical topology, though there can occasionally be more than one hardware variant for the same platform. The feature nodes (telemetry, symptoms, and errors) vary from graph to graph in both structure and value.

\textbf{Homogenization.} To homogenize the heterogeneous inputs (telemetry floats, categorical symptoms, text embeddings), each node type is mapped to a shared $d$-dimensional latent space using type-specific projection heads: MLPs for continuous values and lookup tables for discrete tokens. This aligns disparate feature dimensions before the shared GNN-Transformer layers. The homogenization process creates a mirror graph in which all nodes are converted to a single node set and all edges to a single edge set, mirroring the heterogeneous topology.

\section{Input Features and Component Types}
\label{app:features}

\subsection{Feature Nodes}

Each feature is dynamically connected to the component that generates it in the graph topology; features not generated from a component are connected to the ``software'' node. Each feature becomes a node in the graph with multiple feature values associated with it. For instance, a ``DIMM Error'' feature node has a string feature for the error, a float feature for the maximum error rate, and a float feature for the configured threshold. Table~\ref{tab:feature_list} lists the feature node types. Not every feature is used on every platform, and a small number of features are omitted for IP reasons.

\begin{table}[h]
    \centering
    \small
    \caption{Feature node types and their feature keys.}
    \label{tab:feature_list}
    \begin{tabular}{p{0.27\linewidth} p{0.66\linewidth}}
        \toprule
        \textbf{Feature Node} & \textbf{Feature Keys} \\
        \midrule
        DIMM Error & Error (message / error type); Max Rate (max errors per minute); Threshold (configured error-rate threshold) \\
        CPU Error & Error (message / error type); Max Rate (max errors per minute); Threshold (configured error-rate threshold) \\
        Sensor & Name (sensor identifier); Unit (degrees, volts, RPM, watts, etc.); State (health state: GOOD, DETECTED, MISSING, HOT, COLD, etc.); Violations (number of out-of-range readings) \\
        Component Bandwidth & Flow (connection type: GPU-to-GPU or GPU-to-CPU); Bandwidth (measured value); Threshold \\
        Compute Performance & TFlops; Precision (e.g., FP16, FP32, INT8); Threshold (TFlops threshold) \\
        Symptom & Name (symptom identifier); Test Name (test that produced the symptom); Pass/Fail Status; Message (detailed error or diagnostic output) \\
        Bus Connection Error & Expected / Actual (string reading); Expected / Actual Width; Expected / Actual Speed; Expected / Actual Enabled; Expected / Actual PCIe Type; Expected / Actual Lanes In Use \\
        PCIe Error & Error Type; Max Errors Per Minute; Avg Errors Per Minute; Threshold \\
        In-Band GPU XID Error & Count; PCIe (devpath / endpoint identifier); Link; Fatal (boolean); Error Code (XID numeric code); Error Type; Parallel Test (concurrent test running during error) \\
        Out-of-Band GPU XID Error & Count; Link; Fatal; Error Code (XID numeric code); Error Type \\
        GPU Monitor / BMC Error & Measurement Name; Message Arg (argument extracted from raw Redfish/BMC log); Message (error text) \\
        GPU Count Verification & Count (detected GPUs); Expected Count; Abnormality Degree (difference between expected and actual) \\
        Field Diagnostics & Name; Type; Value (raw stdout output or diagnostic string) \\
        Prior Maintenance Action & Action Type; Preceding Test (test preceding the repair action); Action ID \\
        Prior Historical Symptom & Name; Test Name; Pass/Fail Status; Message \\
        Granular Component Type & Component Type (more detailed component type) \\
        \bottomrule
    \end{tabular}
\end{table}

Symptoms are the most important feature (Table~\ref{tab:feature_importance}) and are emitted as structured results by all diagnostic tests. The data sources for the features are:
\begin{itemize}
    \item All diagnostic tests: Symptom.
    \item Ongoing error monitors: DIMM Error, CPU Error, PCIe Error, Bus Connection Error.
    \item Ongoing sensor monitors: Sensor.
    \item Specific telemetry tests: Component Bandwidth, Compute Performance.
    \item Kernel logs: In-Band GPU XID Error.
    \item BMC/Redfish logs: Out-of-Band GPU XID Error, GPU Monitor / BMC Error.
    \item NVIDIA-specific diagnostic tests: GPU Count Verification.
    \item Standard output of field diagnostics: Field Diagnostics.
    \item Internal databases: Prior Maintenance Action, Prior Historical Symptom.
\end{itemize}

\subsection{Component Nodes}

The component node types are: Accelerator, Accelerator Tray, CPU, CSSD, NIC, DIMM, Fabric Tray, Fan, ICI Cable, Interposer, Motherboard, NVSwitch, PCIe Cable, PCIe Adapter Module, Power Cable, Power Supply, Software, SSD, and Thermal Sensor. Components carry the label information (FRU status, faulty, actions performed, action success). Each component also has a device path, but this feature is deliberately ignored to force the model to learn over the topology structure. The initial representations of component nodes are categorical embeddings of the component type. All sub-FRU components are categorized by the component type of their FRU, but may have different ``Granular Component Type'' features.

\subsection{Feature Importance Ablations}

To further validate the features, we analyzed feature importance on both the TPU v7x and H100 platforms. On TPU v7x, we performed a leave-one-out ablation across the six most common feature node types (Table~\ref{tab:feature_importance}), where lower F1 score, job accuracy, and job recall when a feature is removed indicate higher importance. As expected, symptom features are by far the most critical. These features are derived from hardware diagnostic tests designed by engineers to test the ML accelerators and accompanying hardware. Without them, most failures on TPU v7x cannot be caught, whereas removing any single one of the other five feature types produces metrics within the noise floor of the full model.

\begin{table}[h]
    \centering
    \begin{tabular}{lccc}
        \toprule
        \textbf{Ablated Feature Node Type} & \textbf{Component F1 Score} & \textbf{Job Accuracy} & \textbf{Job Recall} \\
        \midrule
        Full Model (Reference) & 0.9715 & 0.9625 & 0.9764 \\
        \midrule
        symptom & 0.3532 & 0.2426 & 0.2443 \\
        pcie error & 0.9708 & 0.9599 & 0.9695 \\
        sensor & 0.9720 & 0.9634 & 0.9686 \\
        dimm error & 0.9723 & 0.9642 & 0.9808 \\
        bus error & 0.9747 & 0.9625 & 0.9764 \\
        cpu error & 0.9754 & 0.9599 & 0.9808 \\
        \bottomrule
    \end{tabular}
    \vspace{0.5em}
    \caption{Leave-one-out feature ablation on TPU v7x at 10k training steps, ordered from lowest to highest F1 without the feature.}
    \label{tab:feature_importance}
\end{table}

On H100, which has 13 telemetry and diagnostic feature node types, we first evaluate the standalone predictive power of each feature type by training single-feature models that keep only one feature node type at a time and remove the other 12 (Table~\ref{tab:h100_feature_importance}, mean $\pm$ one standard error over three seeds). Keeping only \texttt{symptom} matches the full 13-feature model at convergence ($0.9697 \pm 0.0013$ vs.\ $0.9691 \pm 0.0006$ F1). Among non-symptom signals, broad diagnostic and telemetry node types carry substantial standalone signal, led by \texttt{bus\_error} ($0.5725$ F1), \texttt{field\_diag} ($0.5531$ F1 and $0.6686$ job accuracy), \texttt{sensor} ($0.4654$ F1 and $0.5892$ job accuracy), \texttt{gpu\_count} ($0.3506$ F1), and \texttt{bandwidth} ($0.1780$ F1). Subsystem-specific error counters (\texttt{dimm\_error}, \texttt{xid\_error}, \texttt{oob\_xid\_error}, \texttt{tflops}, \texttt{cpu\_error}, \texttt{pcie\_error}, and \texttt{gpu\_error}) only fire on machines experiencing those specific fault classes and thus yield low overall recall when used in isolation.

\begin{table}[h!]
    \centering
    \setlength{\tabcolsep}{4pt}
    \begin{tabular}{lccc}
        \toprule
        \textbf{Kept Feature Node Type(s)} & \textbf{Component F1 Score} & \textbf{Job Accuracy} & \textbf{Job Recall} \\
        \midrule
        Full model (all 13 features) & $0.9691 \pm 0.0006$ & $0.9688 \pm 0.0002$ & $0.9783 \pm 0.0006$ \\
        All except symptom (12 features) & $0.9270 \pm 0.0060$ & $0.9075 \pm 0.0034$ & $0.9151 \pm 0.0037$ \\
        \midrule
        only symptom & $0.9697 \pm 0.0013$ & $0.9707 \pm 0.0007$ & $0.9791 \pm 0.0006$ \\
        only bus error & $0.5725 \pm 0.0003$ & $0.5031 \pm 0.0002$ & $0.5053 \pm 0.0002$ \\
        only field diag & $0.5531 \pm 0.0030$ & $0.6686 \pm 0.0017$ & $0.6740 \pm 0.0016$ \\
        only sensor & $0.4654 \pm 0.0012$ & $0.5892 \pm 0.0003$ & $0.5909 \pm 0.0009$ \\
        only gpu count & $0.3506 \pm 0.0006$ & $0.4940 \pm 0.0003$ & $0.4954 \pm 0.0005$ \\
        only bandwidth & $0.1780 \pm 0.0895$ & $0.3850 \pm 0.0179$ & $0.4062 \pm 0.0186$ \\
        only dimm error & $0.0595 \pm 0.0000$ & $0.3926 \pm 0.0000$ & $0.3926 \pm 0.0000$ \\
        only xid error & $0.0563 \pm 0.0021$ & $0.3963 \pm 0.0009$ & $0.3990 \pm 0.0009$ \\
        only oob xid error & $0.0094 \pm 0.0009$ & $0.3737 \pm 0.0006$ & $0.3739 \pm 0.0005$ \\
        only tflops & $0.0085 \pm 0.0000$ & $0.3732 \pm 0.0002$ & $0.3734 \pm 0.0000$ \\
        only cpu error & $0.0009 \pm 0.0009$ & $0.3688 \pm 0.0002$ & $0.3690 \pm 0.0000$ \\
        only pcie error & $0.0000 \pm 0.0000$ & $0.3690 \pm 0.0000$ & $0.3690 \pm 0.0000$ \\
        only gpu error & $0.0000 \pm 0.0000$ & $0.3690 \pm 0.0000$ & $0.3690 \pm 0.0000$ \\
        \bottomrule
    \end{tabular}
    \vspace{0.5em}
    \caption{Single-feature inclusion (``keep-only'') ablation on H100 (mean $\pm$ one standard error across 3 seeds at 30k steps). The top two reference rows show the full 13-feature model and the 12-feature model with \texttt{symptom} removed. Each remaining row keeps only the listed feature node type and removes the other 12, ordered by descending F1.}
    \label{tab:h100_feature_importance}
\end{table}

We also performed a leave-one-out ablation across all 13 feature types on H100. Removing \texttt{symptom} reduces component F1 from $0.9691$ to $0.9270$ ($-4.21$ percentage points) and job accuracy from $0.9688$ to $0.9075$ ($-6.13$ percentage points). Unlike TPU v7x, where removing \texttt{symptom} reduces F1 to $0.3532$, combining the 12 non-symptom telemetry, counter, and diagnostic feature types on H100 still recovers over $92\%$ component F1 and $90\%$ job accuracy. Removing \texttt{pcie\_error} produces a small dip to $0.9663$ F1 ($0.9648$ job accuracy), while removing any single one of the remaining feature types leaves F1 within $0.9676$--$0.9699$ and job accuracy within $0.9670$--$0.9715$, confirming that overlapping telemetry and symptom signals compensate when any individual non-symptom feature source is dropped.

\FloatBarrier
\section{Rules-Based Diagnostic System}
\label{app:rules}

The rules-based diagnosis system is the automated diagnostic system used to maintain hardware health prior to \textsc{HardwareDx}, and it is the baseline in our production A/B tests (Section~\ref{sec:experiments}). It operates primarily on a \textit{badness model}, in which components accumulate or lose ``badness'' (a value between 0 and 1) based on incoming signals. As a simple example, if a GPU temperature reading exceeds a threshold, a rule may add a badness of $+0.9$ to the GPU's ``thermal'' failure mode. The production system contains many hundreds of rules of varying complexity.

\subsection{Diagnosis Process}

\textbf{Signal standardization.} The required diagnostic signals are first pulled from health databases and standardized into signals consisting of a reporter, a symptom name, an affected device path, and a timestamp.

\textbf{Rule execution.} The system checks the registry of all rules and activates any that are triggered by the present symptoms. Each FRU component has a number of possible failure modalities (e.g., connectivity, thermal, power). Triggered rules add a badness delta to specific dimensions of one or more components (e.g., increasing the badness score for thermal issues on the CPU). Rules can also have an ``else'' condition, under which a passing test adds a negative badness delta to certain failure modes. More complicated rules can be constructed using compound if-conditions. The badness scores of all triggered rules are aggregated by part-dimension; when multiple rules trigger at the same time (comorbidities), this aggregation determines which failure mode takes priority.

\textbf{Repair selection.} The system identifies the part-dimension pairs with the highest badness score. Based on that failure mode, it looks up a corresponding repair strategy and outputs a sequence of actions. These are compared against prior actions, with post-processing to avoid repeating actions and to improve efficacy.

\subsection{Types of Rules}

There are three general classes of rules that can be added to the rules registry:
\begin{itemize}
    \item \textbf{Standard rules:} If the reporter (e.g., a PCIe connectivity test) and symptom (e.g., low throughput) match, the rule applies the specified badness deltas.
    \item \textbf{Rate-limited rules:} Identical to standard rules, but triggered at most once per device path per iteration. This is useful for repetitive reporters or multiple test loops, preventing artificial badness inflation.
    \item \textbf{High-confidence overrides:} For highly unambiguous signals, the rule directly triggers a specific repair strategy.
\end{itemize}
In addition to these configurable rules, certain rules are written directly in code to handle more complex logic.

\subsection{Development and Maintenance of Rules}

Every rule is onboarded via direct engineering effort. Because the rules are mostly config-based, engineers from new-platform onboarding teams as well as from external teams and vendors are able to onboard rules effectively. While updating the rules is straightforward, discovering and determining which rules to add is a highly involved process requiring significant time and engineering effort. Badness scores can be initially determined by hardware and diagnostic engineers, but they require ongoing refinement as data is collected and performance gaps are found.

\subsection{Rules Baseline in Longitudinal Comparisons}
\label{app:rules_baseline}

Table~\ref{tab:performance_timeline} evaluates the models that had been developed as of HW-GA and as of HW-GA+4mo, re-evaluated on a unified held-out dataset for consistent comparison. Due to the complexity of the rules-based system, there is no way to evaluate historical versions of the system as a whole. Instead, we use the rules' recommendations at the time of the actual hardware repair as the Rules@HW-GA baseline. Because the rules are continuously updated in production by a large engineering team, this baseline already includes ongoing rule enhancements. This biases the comparison in favor of the rules-based system, making the reported improvements of \textsc{HardwareDx} conservative.

\section{Production A/B Evaluation Details}
\label{app:ab_details}

\textbf{Baseline.} The A/B tests compare \textsc{HardwareDx} against the deployed production diagnostic system, i.e., the rules-based system described in Appendix~\ref{app:rules}. This system sees heavy, ongoing engineering investment (greater than that of \textsc{HardwareDx}) and is a highly competitive baseline.

\textbf{Assignment.} Traffic is split at the machine level between the production rules-based system and \textsc{HardwareDx}. We use 20\% of machines for the TPU platform and 50\% of machines for the H100 pilot, which has lower repair volume. The analysis covers 225 H100 repair windows and 2,567 TPU repair windows.

\textbf{Metrics.} The reduction in downtime (mean TTR) is computed from the mean time a machine spends in the test, diagnosis, and repair loop before returning to serving; install time is excluded. The QA delivery metric is based on contractual service level objectives: to turn over a cluster of machines, a certain percentage of compute nodes must be fully online and functioning. We compute the percentile of repair cases that must be completed to reach that service level objective and report the repair time at that percentile.

\textbf{Uncertainty.} Standard errors of the mean TTR are computed with classical parametric estimation, $\mathrm{SE} = s/\sqrt{n}$, where $s$ is the sample standard deviation and $n$ the number of repair windows. Standard errors for the QA delivery reduction, which is a percentile, are computed non-parametrically using the Woodruff method \citep{woodruff1952confidence}, based on the binomial distribution of order statistics. For the H100 fleet, \textsc{HardwareDx} reduced mean TTR by 18.3\%$\pm$7.6\% and QA delivery time by 28.3\%$\pm$10.7\%. For the TPU platform, it reduced mean TTR by 16.1\%$\pm$2.6\% and QA delivery time by 25.1\%$\pm$2.5\% ($\pm$ one standard error). Even though the H100 analysis is based on fewer repair windows, the improvements remain statistically significant.

\section{Data Collection}
\label{app:data_labeling}

For supervised data collection, we utilize expert recommendations. For each job being labeled, the expert will look at the full case history and the repair-ability guidelines for the failing diagnostic tests before selecting which components should be touched. This is a highly technical and involved process. Table \ref{tab:dataset_stats} shows the collected dataset volumes.

\begin{table}[]
    \centering
    \caption{HardwareDx Component Training Dataset Statistics}
    \label{tab:dataset_stats}
    \begin{tabular}{lrrrr}
        \toprule
        \textbf{Platform} & \textbf{Unsupervised Dataset} & \textbf{Labeled} & \textbf{Median Graph} & \textbf{Unique} \\
        & \textbf{Size (Trajectories)} & \textbf{Cases} & \textbf{Size (Nodes)} & \textbf{Labels} \\
        \midrule
        TPU v5p         & 111,764 & 5,236 & 874   & 59 \\
        TPU v7x         & 43,987  & 7,529 & 992   & 66 \\
        TPU v7          & 1,381   & 245    & 1,142 & 29 \\
        H100            & 81,015  & 6,780 & 1,613 & 85 \\
        H200            & 7,558   & 1,838   & 1,276 & 80 \\
        B200            & 18,481  & 1,667   & 1,379 & 91 \\
        \bottomrule
    \end{tabular}
\end{table}

\subsection{Dataset characteristics and Topology}
Table \ref{tab:dataset_stats} provides the primary training datasets utilized by \textsc{HardwareDx}. TPU v5p and H100 represent our most data-rich environments, each containing over 111,000 and 81,000 trajectories, respectively. The H100 platform shows the most complex physical topology, requiring a median of 1,613 nodes per machine snapshot to represent all components, subcomponents, and feature signals. 

\subsection{Label Diversity}
The B200 and H100 platforms show the greatest variety of unique labels (91 and 85, respectively), which indicates a broad capture of failure modes for these architectures. Meanwhile, the H100 platform has the highest number of labeled windows. TPU v7x similarly maintains a highly robust set of ground truth-examples despite having a smaller overall dataset size compared to TPU v5p and H100.

\section{Success-Rate Inference}
\label{app:succ_rate_inference}

As mentioned in the paper, the success-rate inference strategy aims to enable \textsc{HardwareDx} to directly predict actions based on unsupervised learning. This strategy aims to take the most likely to succeed actions while avoiding unreasonable actions (those that would never be seen in normal data center operations), while also avoiding taking excessive actions, even if reasonable. Actions are chosen based on (1) the component needing an action applied, (2) the action being reasonable to expect, and (3) the action having higher chance of success than any other action for the component. 

\subsection{No-Action Gating} 
Let $a_{\text{na}} \in \mathcal{A}$ denote the ``no-action'' operation, and let $\hat{P}_{\text{succ}}^{(c,\text{na})}$ represent the predicted conditional likelihood of the machine passing the next test if no physical repair is performed on component $c$. The inference procedure follows a three-step decision process:

For each component, we first evaluate whether physical intervention is strictly necessary. If the predicted likelihood of success without intervention exceeds a predefined threshold $\tau_{\text{na}}$, the system recommends no action for that component. This prevents unnecessary hardware swaps in the presence of transient faults (e.g., software timeouts that resolve upon reboot).
\begin{equation}
    \text{If } \hat{P}_{\text{succ}}^{(c,\text{na})} \ge \tau_{\text{na}}, \text{ return } a_{\text{na}}
\end{equation}

\subsection{Propensity Filtering} 
For each component, $c \in \mathcal{C}$, where intervention is required ($\hat{P}_{\text{succ}}^{(c,\text{na})} < \tau_{\text{na}}$), we define a candidate set of component-action pairs, $\mathcal{V}$, by filtering the action space using the behavioral prior learned in Task 2. We only consider actions whose predicted propensity exceeds a threshold $\tau_{p}$:
\begin{equation}
    \mathcal{V}_c = \left\{ a \in \left( \mathcal{A} \setminus \{a_{\text{na}}\} \right) \;\middle|\; \hat{P}_{\text{prop}}^{\left(c,a\right)} \ge \tau_{p} \right\}
\end{equation}

This bounds our search space to the manifold of historically reasonable technician behaviors, preventing the model from prescribing out-of-distribution or structurally invalid actions (following \citet{kumar2019stabilizing}). The mathematical gating mechanism ($\tau_p$) effectively bounds the model's policy within the manifold of physically safe, historically validated technician behaviors, prioritizing hardware safety over aggressive exploration.

\subsection{Maximum Likelihood Selection} 
Finally, from the filtered candidate set $\mathcal{V}$, we select the action $a^*$ for component $c$ that maximize the conditional likelihood of success, as predicted by Task 3:
\begin{equation}
    a^* = \underset{a \in \mathcal{V}_c}{\arg\max} \;\hat{P}_{\text{succ}}^{(c,a)}
\end{equation}

If $\mathcal{V}_c$ is empty, we fallback to $a_{\text{na}}$.

By chaining these heads, \textsc{HardwareDx} balances aggressive automated repairs with historical safety priors, ensuring that the prescribed actions are both highly likely to succeed and operationally sound.

\section{Simulated Evaluation of Success-Rate Inference}
\label{app:sim_eval_succ_rate}

We conducted additional evaluation on the Success-Rate Inference strategy for the unsupervised tasks using a window-based simulation approach. We collect a set of 941 recent repair trajectories of the form 

\begin{equation}
    \tau = (x_0, \dot{A}_0, x_1, \dot{A}_1, \dots, x_{T-1}, \dot{A}_{T-1}, x_T)
\end{equation}

where $x_0, ..., x_{T-1}$ are the failing tests, $x_T$ is the final passing test, and each $\dot{A}_i\in \left(\mathcal{C}\times\mathcal{A}/a_{\text{na}}\right)^*$ is the set of repair actions actually performed at that iteration, excluding "no-action". This is a slight redefinition of notation from what was used in Section~\ref{sec:model}, but represents the same underlying data. While we cannot know the utility of every action in the trajectory, we can assume that actions $\dot{A}_{T-1}$ are likely to be useful as they led to the final tests passing. To cover a wide range of situations we sample the windows based on the presenting symptoms and the number of iterations it took to fix. This makes the selected windows more varied and more difficult than the natural distribution. 

We cannot simulate the effect of running \textsc{HardwareDx}'s success-based inference from the start of the window as the observed test results and simulated test results would diverge as soon as the actions diverge. However, we can simulate the effect of running \textsc{HardwareDx} independently on each failing test $x_0$ through $x_{T-1}$. Let $\hat{A}_0, ..., \hat{A}_{T-1}$ be the \textsc{HardwareDx} suggested sets of actions. 

We then evaluate how often \textsc{HardwareDx} replicates the final actions $\dot{A}_{T-1}$ in its prior predictions:

\begin{equation}
    \text{Final Action Replication Rate} = \mathbb{E}_{\tau}\left[\dot{A}_{T-1} \subseteq \left(\hat{A}_0 \cup ... \cup \hat{A}_{T-1} \right)\right]
\end{equation}

We also measure an action-weighted version. This is meant to account for the fact that not every one of the final actions is expected to be necessary. For instance, if there were three parts swapped, potentially only one of those swaps was necessary. This metric gives the model partial credit for replicating a subset of the final actions. 

\begin{equation}
    \text{Weighted Final Action Replication Rate} = \mathbb{E}_{\tau}\left[\frac{1}{|\dot{A}_{T-1}|}\sum_{a\in \dot{A}_{T-1}}\left[a \in \left(\hat{A}_0 \cup ... \cup \hat{A}_{T-1} \right)\right]\right]
\end{equation}

Of the samples that were replicated, we can also measure the estimated iterations it takes to reproduce the final actions. 

\begin{equation}
\text{Estimated Iterations} = 1+\min \left\{ i \in {0, \dots, T-1} \mid \dot{A}_{T-1} \subseteq \bigcup_{j=0}^{i} \hat{A}_j \right\}
\end{equation}

Table \ref{tab:drwatson_window_based_comparison} shows the results of this window based evaluation. The \textsc{HardwareDx} approach shows a better final action replication rate among these windows while also offering a better estimated improvement in iterations. 

\begin{table}[]
    \centering
    \caption{Performance Comparison: Existing Automated System vs. \textsc{HardwareDx} Predict Repair Actions with Success Rate}
    \label{tab:drwatson_window_based_comparison}
    \begin{tabular}{lcc}
        \toprule
        \textbf{Metric} & 
        \textbf{\begin{tabular}{@{}c@{}}Existing \\ System\end{tabular}} & 
        \textbf{\begin{tabular}{@{}c@{}}HardwareDx Predict \\ Repair Actions\end{tabular}} \\
        \midrule
        Final Action Replication Rate               & 56.23\% & \textbf{63.42\%} \\
        Action-Weighted Final Action Replication Rate & 62.50\% & \textbf{73.04\%} \\
        Average Estimated Iterations ($\Delta$ over actual iterations)      & -0.29 & \textbf{-0.50} \\
        \bottomrule
    \end{tabular}
\end{table}

\section{Finetuning Two-Platform Model}
\label{app:complete_fm_results}

We run an additional experiment in which we train a two-platform foundation model on H100 and H200 data before fine-tuning on complete B200 data. Table~\ref{tab:2platform_finetune} shows these results. Using the foundation model as a starting point improves per component F1-score and job-level accuracy, demonstrating the generalization capability. 

\begin{table}[]
    \centering
    \caption{Results of finetuning a foundation model pretrained on H100 and H200 data on the B200 platform. }
    \label{tab:2platform_finetune}
    \begin{tabular}{l ccc}
        \toprule
        & \multicolumn{3}{c}{\textbf{B200 (Finetune Platform)}} \\
        \cmidrule(l){2-4}
        
        \textbf{Model} & Job-level Acc & Job-level Rec & Component F1 \\
        \midrule
        
        \quad FM-2Platform-FT  & \textbf{0.9848} & \textbf{0.9939} & \textbf{0.9817} \\
        \quad Single           & 0.9787          & \textbf{0.9939} & 0.9728 \\
        \bottomrule
    \end{tabular}
\end{table}

\section{Additional Ablations}
\label{app:additional_ablations}

\begin{figure}[]
     \centering
     \begin{subfigure}[b]{0.48\textwidth}
         \centering
         \includegraphics[width=\linewidth]{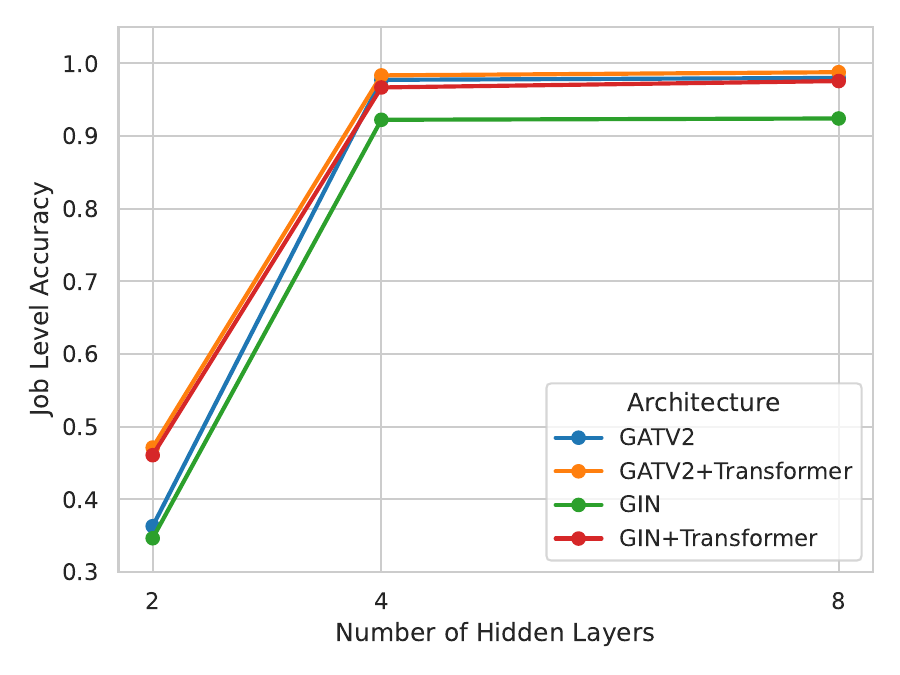}
         \caption{Job-Level Accuracy}
         \label{fig:layer_acc}
     \end{subfigure}
     \hfill
     \begin{subfigure}[b]{0.48\textwidth}
         \centering
         \includegraphics[width=\linewidth]{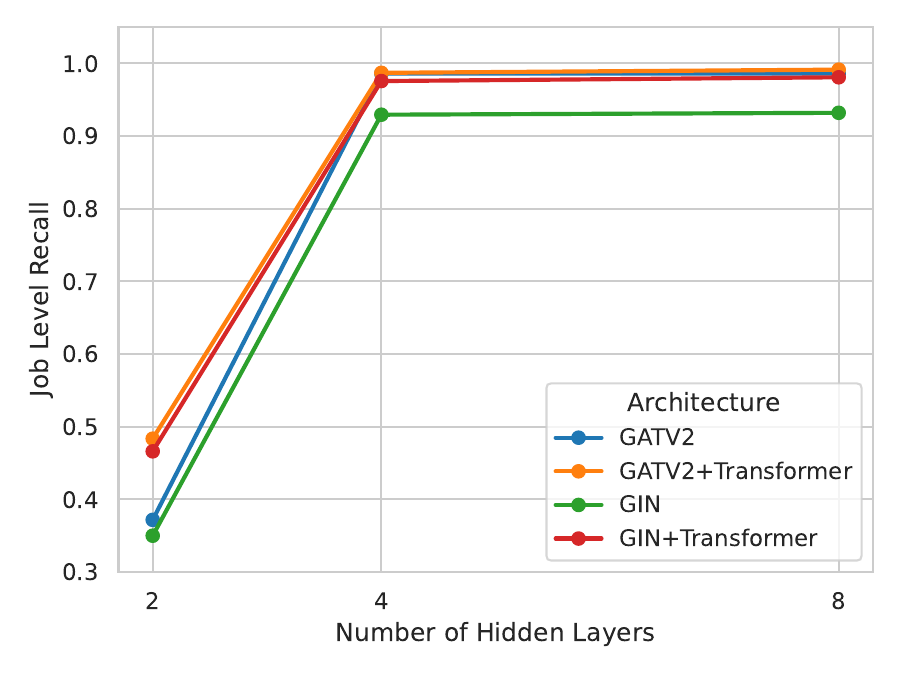}
         \caption{Job-Level Recall}
         \label{fig:layer_recall}
     \end{subfigure}
     
     \vspace{0.3cm} 
     
     \begin{subfigure}[b]{0.48\textwidth}
         \centering
         \includegraphics[width=\linewidth]{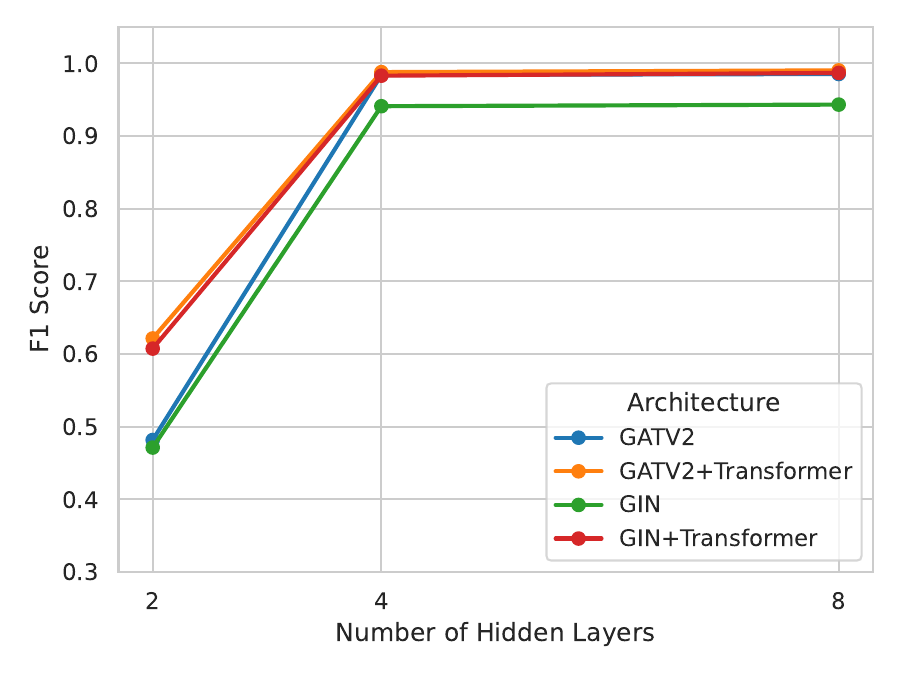}
         \caption{F1-Score}
         \label{fig:layer_gen}
     \end{subfigure}
     
     \caption{Comprehensive ablation study on the number of layers.}
     \label{fig:layer_ablation_group}
\end{figure}

We present additional ablation experiments here. Table \ref{tab:ablation_gnn} in the main paper presented the primary ablation on core GNN choice. Figure~\ref{fig:layer_ablation_group} shows the results of tuning number of layers. Two layers is clearly not enough to adequately learn the data. While performance does continue to improve, the discrepancy between four and eight layers is much smaller. Interestingly, with only two layers, both GATv2 and GIN perform far worse than with the Transformer included, but with more layers, GATv2 only performs slightly worse while a large gap remains to GIN. 

\begin{table}[]
    \centering
    \caption{GNN Model Calibration on TPU v7x using Expected Calibration Error and Brier Scores (lower is better). The evaluated models both include transformer layers but include different local GNN layers.}
    \label{tab:calibration_results}
    \begin{tabular}{lcc}
        \toprule
        \textbf{\begin{tabular}{@{}l@{}}GNN \\ Layer\end{tabular}} & 
        \textbf{\begin{tabular}{@{}c@{}}Expected Calibration \\ Error ($\downarrow$)\end{tabular}} & 
        \textbf{\begin{tabular}{@{}c@{}}Brier \\ Score ($\downarrow$)\end{tabular}} \\
        \midrule
        GATv2 & 0.0015 & 0.0062 \\
        GIN   & 0.0020 & 0.0067 \\
        \bottomrule
    \end{tabular}
\end{table}

We further present some additional evaluation on calibration metrics for the Task 2 and 3 predictions. We present the Expected Calibration Error and Brier Score of our foundation model in table \ref{tab:calibration_results}. Expected Calibration Error is a pure calibration measure: it measures how close the model predicted rate is to the true rate (e.g. of an action being taken or of the next test passing) across different bins \citep{Naeini2015ObtainingWC}. However, this can be gamed by predicting more conservatively, such as only predicting the average rate of incidence. Brier Score measures calibration but also factors in the discriminative power of the predictions \citep{Brier1950VERIFICATIONOF}. 

\section{Hyperparameters}
\label{app:hyperparams}

We conduct hyperparameter sweeps across learning rates of 2e-3, 2e-4, 2e-5, 2e-6 using a cosine decay schedule with 1000 warmup steps. We evaluate per-task batch sizes of [8, 16] (measured in number of graphs per batch). We evaluate latent dimensions of [128, 256]. We evaluate layer depth of [2, 4, 8]. Models are trained with AdamW using $\beta_1=0.9$ and $\beta_2=0.999$. Foundation model training is done for 50,000 steps (with occasional early stopping). Fine-tuning is performed for 10,000 steps. We split data 80\% for training and 20\% for evaluation. Prior to production, we confirm the model functions on a small held out set of recent unseen data to provide further validation. 


\end{document}